\documentclass[10pt,journal,compsoc]{IEEEtran}

\usepackage{microtype}
\usepackage{graphicx}
\usepackage{subcaption}
\usepackage{booktabs} 
\ifCLASSOPTIONcompsoc
  \usepackage[nocompress]{cite}
\else
  \usepackage{cite}
\fi

\usepackage{amsmath,amsfonts, amsthm, amssymb}
\usepackage{mathtools}
\usepackage{xcolor}
\usepackage{glossaries}
\usepackage{tabularx}
\usepackage{authblk}
\usepackage{algorithm}
\usepackage[noend]{algpseudocode}

\usepackage[utf8]{inputenc}
\usepackage{pgfplots}
\pgfplotsset{compat=newest}
\usepgfplotslibrary{groupplots}
\usepgfplotslibrary{dateplot}
\usepackage{hyperref}
\usepackage{stackengine}
\def\delequal{\mathrel{\ensurestackMath{\stackon[1pt]{=}{\scriptscriptstyle\Delta}}}}

\makeatletter
\newcommand{\pushright}[1]{\ifmeasuring@#1\else\omit\hfill$\displaystyle#1$\fi\ignorespaces}
\makeatother

\DeclarePairedDelimiter\abs{\lvert}{\rvert}%
\DeclarePairedDelimiter\norm{\lVert}{\rVert}%

\makeatletter
\let\oldabs\abs
\def\abs{\@ifstar{\oldabs}{\oldabs*}}
\let\oldnorm\norm
\def\norm{\@ifstar{\oldnorm}{\oldnorm*}}
\makeatother

\DeclarePairedDelimiterX{\KLD}[2]{(}{)}{%
  #1\;\delimsize\|\;#2%
}

\def\KL{\mathrm{KL}}
\def\HH{{\cal H}}

\def\EE{{\rm I\hspace{-0.50ex}E}}
\newcommand{\EEc}[1]{{\EE}_{#1}}
\newcommand{\Objective}{\mathcal{J}}
\newcommand{\statespace}{\mathcal{S}}
\newcommand{\actionspace}{\mathcal{A}}
\newcommand{\Rmodel}{\mathcal{R}}
\newcommand{\Pmodel}{\mathcal{P}}
\newcommand{\Imodel}{\mathcal{\iota}}
\newcommand{\KLM}{\KL\KLD}

\DeclareMathOperator*{\argmax}{arg\,max}

\newacronym{rl}{RL}{Reinforcement Learning}
\newacronym{drl}{DRL}{Deep Reinforcement Learning}
\newacronym{hrl}{HRL}{Hierarchical Reinforcement Learning}
\newacronym{avi}{AVI}{Approximate Value-Iteration}
\newacronym{api}{API}{Approximate Policy-Iteration}
\newacronym[plural=MDPs, firstplural=Markov Decision Processes (MDPs)]{mdp}{MDP}{Markov Decision Process}
\newacronym{kl}{KL}{Kullback-Leibler Divergence}
\newacronym{gae}{GAE}{Generalized Advantage Estimation}
\newacronym{papi}{PAPI}{Projections for Approximate Policy Iteration}
\newacronym{her}{HER}{Hindsight Experience Replay}
\newacronym{ham}{HAM}{Hierarchy of Abstract Machines}
\def\sizechetah{.2}

\begin{document}
\title{Continuous Action Reinforcement Learning from a Mixture of Interpretable Experts}

%

\author{Riad~Akrour,
        Davide~Tateo,
        and~Jan~Peters%
\thanks{Code: \url{https://github.com/akrouriad/tpami_metricrl}. R. Akrour is with Aalto University, Finland. D. Tateo and J. Peters are with TU Darmstadt, Germany.}}         


\markboth{ }%
{Shell \MakeLowercase{\textit{et al.}}: Reinforcement Learning from a Mixture of Interpretable Experts}

\IEEEpubid{}

\IEEEtitleabstractindextext{%
\begin{abstract}
	Reinforcement learning (RL) has demonstrated its ability to solve high dimensional tasks by leveraging non-linear function approximators. However, these successes are mostly achieved by 'black-box' policies in simulated domains. When deploying RL to the real world, several concerns regarding the use of a 'black-box' policy might be raised. In order to make the learned policies more transparent, we propose in this paper a policy iteration scheme that retains a complex function approximator for its internal value predictions but constrains the policy to have a concise, hierarchical, and human-readable structure, based on a mixture of interpretable experts. Each expert selects a primitive action according to a distance to a prototypical state. A key design decision to keep such experts interpretable is to select the prototypical states from trajectory data. The main technical contribution of the paper is to address the challenges introduced by this non-differentiable prototypical state selection procedure. Experimentally, we show that our proposed algorithm can learn compelling policies on continuous action deep RL benchmarks, matching the performance of neural network based policies, but returning policies that are more amenable to human inspection than neural network or linear-in-feature policies.
\end{abstract}

}

\maketitle

\IEEEdisplaynontitleabstractindextext

\IEEEpeerreviewmaketitle

\IEEEraisesectionheading{\section{Introduction}}

\gls{rl} \cite{Sutton98,Szepesvari10} has led to several practical breakthroughs despite the high dimensionality of the state-action space of the problems at hand~\cite{Mnih15,Silver16}. To do so, recent \gls{rl} algorithms learn complex function approximators, typically 'black-box' deep neural networks~\cite{Mnih15,lillicrap15}. However, the opacity of these function approximators might be a cause for concern. Because of this opacity, simpler and more interpretable policies have been suggested for robotics in the past for specific applications~\cite{marjaninejad2019autonomous,sundaresan2020learning}. In this paper, we propose a general purpose learning algorithm that returns a human-readable policy, facilitating the analysis of the learned behavior by making it more transparent to the human. A human-readable policy can help the human understanding and learning the task. For instance, in the Rubik's cube puzzle, while it might be challenging to some, one can consult publicly available policies when stuck, find the state in which they are and take the corresponding action to progress. We propose in this paper an \gls{rl} algorithm that returns a mixture of expert policy, based on similarity with prototypical states as with the Rubik's cube policy. From a learning theory point of view, a human-readable policy will be simpler and hence policies returned by our algorithm might generalize better. 

\gls{rl} algorithms use function approximators for estimating the value function~\cite{Mnih15,van2016deep} or both the value function and the policy~\cite{mnih16,lillicrap15, Schulman15}. For the sake of interpretability, we propose to replace the policy, with a more interpretable structure. We choose to use an interpretable structure for the policy instead of the value function because the policy usually has a simpler functional shape \cite{Rexakis08,Anderson00}, and can thus be more easily approximated. Our proposed algorithm yields a clustering of the state space, where a single action is associated with each cluster. Such clustering of the state space could be seen as a form of state  abstraction~\cite{Li06,Abel16,Mankowitz16,Akrour18}. Previous research in high-dimensional state abstraction however, only considered non-interpretable state-to-cluster mappings, in the form of polynomial functions \cite{Mankowitz16} or neural networks~\cite{Akrour18}.

As a computer program, the policy returned by our algorithm can be seen as a sequence of \texttt{IF} blocks having the structure \texttt{IF close(state, center[k]) DO action[k]}; where \texttt{state} is the current state, \texttt{center[k]} is a cluster center and \texttt{action[k]} its associated action. To ensure that such a policy is interpretable, we impose a series of limitations that make the underlying optimization problem challenging. First, we limit beforehand the number of clusters to a fixed (small) number K. Second, and most importantly, we do not allow the cluster centers to be optimized by gradient descent. Instead, we only allow cluster centers to be picked from the stream of states encountered during learning. Doing so ensures that the cluster centers are within the potentially lower-dimensional manifold that is the state space and hence, are interpretable. In the experiments section we show that when the decision of hand-picking the prototypical states from trajectory data is bypassed, and prototypes are instead optimized by gradient descent, the resulting policy is significantly harder to interpret and the relation between the observed trajectory and the experts is not obvious.

The final component that is critical for interpretability in our policy structure is the function \texttt{close} that discriminates if a state belongs to the specified cluster, i.e., the state is sufficiently close to the cluster prototype. In this paper, we assume that this function is based on the Euclidean distance. We leave the problem of \textbf{learning} an appropriate metric for more complex input spaces such as images to future work.


\section{State-of-the-art}
In this section we review prior research on two aspects of reinforcement learning (RL). As a mixture of experts our algorithm is closely related to hierarchical RL and state abstraction, and we review this topic in the first subsection. Since we desire the learned experts to be interpretable, we review in the second subsection the topic of interpretable RL. 
\subsection{Hierarchical reinforcement learning}
Abstraction in RL has long been seen as a promising direction to combat the curse of dimensionality by exploiting the structure of the problem at hand \cite{Precup98, Sutton99, Dietterich99}. We distinguish in this section two cases: temporal abstraction \cite{Konidaris09, Silva12, Hausknecht15, Daniel16, Bacon17} and state abstraction \cite{Li06, Abel16, Mankowitz16}. Temporal abstraction divides a complex task into temporally extended sub-policies, often called lower-level policies or skills and is commonly referred to as \gls{hrl}. The objectives of \gls{hrl} approaches are: (i) to have a more structured exploration (ii) to exploit structural expert knowledge (iii) to increase interpretability of the learned behaviors. Many different \gls{hrl} frameworks exist~\cite{barto2003recent}, some put focus on exploiting expert prior knowledge~\cite{Parr1997ReinforcementLW,Dietterich2000HierarchicalRL,tateo2019graph}, others put more focus on providing structured exploration~\cite{Sutton1999BetweenMA}.

In recent years, the focus has shifted away from expert knowledge and interpretability, to the applicability of \gls{hrl} to harder tasks and to learning efficiency, thanks to the use of deep RL techniques.
Some approaches extend the Feudal Q-Learning algorithm~\cite{dayan1993feudal}, such as~\cite{vezhnevets2017feudal, nachum2018data}. Others are based on the Option framework~\cite{Sutton1999BetweenMA} such as~\cite{bacon2017option,riemer2020role}. 
Another class of \gls{hrl} methods are based on \gls{her}~\cite{andrychowicz2017hindsight}, such as~\cite{levy2018learning}. Finally, other methods are based on pre-training the lower-level skills~\cite{florensa2017stochastic, co2018self, tessler2017deep,eysenbach2018diversity}.
All these approaches have in common the fact that the obtained policies and skills are difficult to interpret due to the use of neural network policy architectures. We show in this paper that on many continuous action deep RL benchmarks, a simpler and more interpretable policy structure can still be competitive with its neural network counterpart. 

While the policy we present is hierarchical in nature, it does not make use of temporally extended actions, and abstraction is rather at the level of the state space. In the context of bandit algorithms, our work is related to clustered linear bandits \cite{gentile14, korda16, mahadik20} where the goal is to uncover both a clustering of users and an associated action to each cluster maximizing the rewards. However, we cannot rely in our case on the assumption of linearity of reward functions as in linear bandits. In the context of RL, our approach is close to the topic of state abstraction, where one exploits the structure of a Markov Decision Process (MDP) to derive an \textit{abstract} MDP having a compressed state space but a similar optimal policy to that of the ground MDP. Structure in state abstraction emerges by grouping together several states from the ground MDP into an abstract state. Several criteria were studied to group states together in the exact case where the optimal policies of the abstract and ground MDPs coincide \cite{Li06} and the approximate case \cite{Abel16} where the optimal policy of the abstract MDP is only near-optimal in the ground MDP. However, all these criteria require knowledge of the transition function. In this paper we consider instead a more realistic scenario where the transition function is unknown, and propose clustering criteria that try to best cover trajectory data generated throughout the learning process.


\subsection{Interpretable reinforcement learning}
There are two main ways of achieving interpretability in supervised learning:  i) learning a black-box model and making it interpretable post-hoc by sensitivity analysis or by (locally) mimicking the black-box with an interpretable model~\cite{Strumbelj10, Ribeiro16, Craven96} or ii) by learning an interpretable model from the get-go such as linear models, decision trees or attention-based networks~\cite{Ustun15, Letham15, Vaswani17}. In RL, a similar categorization exists, where prior publications either considered explaining learned policies~\cite{Madumal19, Hayes17}, learning an interpretable policy by imitation learning of a neural network policy~\cite{Verma18, Bastani18} or learning an interpretable policy from scratch using for example linear policies~\cite{Nie19}. 

However, we believe that the choice of learning an interpretable structure from scratch is more appropriate than making it interpretable post-hoc, due to the intrinsic sequential nature of \gls{rl}. First, there is an inherent difficulty in imitating a complex policy with a simpler model, and the unavoidable differences will compound quadratically w.r.t. the problem's horizon, due to a drift in state distributions~\cite{ross10}. This drift in state distributions renders a local interpretation---i.e. an interpretation w.r.t. a given state---less informative than in supervised learning where the input data distribution is unaffected by the change to a simpler model. Finally, given the fixed complexity of an interpretable policy, typically lower than that of a neural network, the policy might need to adopt a vastly different strategy---e.g. crawling vs running---to the problem. By learning an interpretable policy with RL, we ensure that an optimal policy w.r.t. the particular interpretable policy class is within reach of the learning algorithm. 

Decision trees are interpretable structures, and there has been prior research on learning decision tree policies with \gls{rl} such as with the Fitted Q Iteration (FQI) algorithm~\cite{Ernst05}. However, learning small enough trees to be interpretable was only demonstrated using imitation learning \cite{Bastani18, Coppens19}, where a significant reduction in tree size was obtained compared to FQI~\cite{Bastani18}. As for linear policies, it is important to note that by linear policies we mean linear-in-state as opposed to linear-in-feature, which might perform better~\cite{rajeswaran17}, but introduces complex transformations of the state space, hindering interpretability. An example of an interpretable linear policy is in the article of~\cite{Nie19}, where a medical treatment policy was learned. The policy was easy to observe globally as the state was two dimensional. But linear-in-state policies are not always an ideal choice for human-readable policies. First, their interpretability in higher dimensional spaces can be questionable especially when they are not sparse. Secondly, a linear policy class might severely limit the quality of the policy. In this paper, we explore an alternative policy structure akin to nearest neighbors and (Gaussian) kernel-based models. The policy complexity can be increased to match the complexity of the problem but unlike decision trees, part of the policy is differentiable and easier to train.

Kernel-based \gls{rl} methods have already been explored in the past, such as in the article of \cite{Xu07} that is especially related to ours. Specifically, \cite{Xu07} developed a kernel-based extension of least squared policy iteration~\cite{lagoudakis03} and added a compression step to limit the complexity of the policy. Keeping the complexity low is key to providing a human-readable policy. However, in their experiments, more than 3000 centers were used for a double pendulum task with a 4-dimensional state space. In contrast, we are able to learn a walking gait on the \texttt{AntBulletEnv-v0} environment, that has a 28-dimensional state space, with as little as 10 centers. To the best of our knowledge, no other kernel-based \gls{rl} approach has achieved this level of policy simplicity on a task of this complexity.

\section{Preliminaries}
We introduce in this section elementary notions of reinforcement learning and specific optimization tools that we will use for tackling the learning problem of our mixture of interpretable experts as framed in Sec.~\ref{sec:opt}.
\subsection{Approximate policy iteration}
Our algorithm is couched in the Approximate Policy Iteration~(API) framework \cite{Bertsekas11,Scherrer14} which requires the introduction of a few notations. First, a \gls{mdp} is defined as a $6$-tuple $\mathcal{M} = <\statespace, \actionspace, \Pmodel, \Rmodel, \gamma, \Imodel>$, where $\statespace$ is the state space, $\actionspace$ is the action space, assumed to be both continuous spaces, $\Pmodel:\statespace \times \actionspace \times \statespace \to \mathbb{R}$ is the transition distribution where $\mathcal{P}(s' | s, a)$ is the probability density of reaching state $s'$ when performing action $a$ in state $s$, $\mathcal{R}: \mathcal{S} \times \mathcal{A} \to \mathbb{R}$ is the reward function, $\gamma \in (0,1]$ is the discount factor, and $\Imodel: \statespace \to \mathbb{R}$ is the initial state distribution. A stochastic policy $\pi: \mathcal{S} \times{A} \to \mathbb{R}$ is a function such that the probability density of taking action $a$ in state $s$ is $\pi(a|s)$. Let $Q^{\pi}(s,a) = \EEc{s_t,a_t\sim\mathcal{P},\pi}\left[\sum_{t=0}^\infty\gamma^t R(s_t, a_t)\mid s_0  = s, a_0=a\right]$ be the the Q function of policy $\pi$. Let $V_{\pi}(s) = \EEc{a\sim\pi(.|s)}\left[Q_{\pi}(s,a)\right]$ and $A_{\pi}(s,a) = Q_{\pi}(s,a) - V_{\pi}(s)$ be, respectively, the value and the advantage function of $\pi$.

A parametric policy is a family of policies $\pi_\theta$ such that the probability distribution depends on a vector of parameters $\theta$. Let $\Objective(\pi_\theta)=\EEc{s_0\sim{\Imodel}}\left[V_{\pi_\theta}(s_0)\right]$ be the performance of the parametric policy $\pi_\theta$ and $\pi_{\theta^*}=\argmax_{\pi_\theta}\Objective(\pi_\theta)$ be the optimal policy.
To find the optimal policy  $\pi_{\theta^*}$ using API, we sample at each iteration trajectories, evaluate the policy and update the policy parameters. The presentation of the paper focuses on the policy update step, since the sampling and policy evaluation steps use instead standard techniques. Specifically, the advantage function of the current policy is evaluated on the generated samples by learning a neural approximated value function, following standard procedures described in e.g. \cite{schulman16}.
\subsection{Projections for API}
\begin{algorithm}[t]
   \caption{Linear-Gaussian policy projection for KL and Entropy constraints}
   \label{alg:linproj}

    \begin{algorithmic}[1]
        \Require 
        \Statex$\pi_{\theta}(.|s)\delequal\mathcal{N}(M\psi(s), \Sigma)$, where $\theta\delequal\lbrace M, \Sigma\rbrace$
        \Statex $q(.|s)\delequal\mathcal{N}(M_q\psi_q(s), \Sigma_q)$
        \Statex $\EEc{s\in\mathcal{T}}\left[\KLM{\mathcal{N}(M_q\psi(s), \Sigma_q)}{\mathcal{N}(M_q\psi_q(s), \Sigma_q)}\right] < \epsilon$
        \vspace{0.2cm}
        \Procedure{computeProjection}{$\pi_\theta, q, \mathcal{T}$}
           \State $\Sigma\leftarrow$\Call{Entropy\_projection}{$\Sigma$}\label{alg:line:entropy}
           \vspace{0.2cm}
           \If {$\EE_s\left[\KLM{\mathcal{N}(M_q\psi(s), \Sigma)}{q(.|s)}\right] > \epsilon$}\label{alg:line:check1}
        	  \State $\eta_g \leftarrow \dfrac{\epsilon - m_q(M_q)}{m_q(M)+r_q(\Sigma)+e_q(\Sigma)}$ \Comment{$m_q, r_q, e_q$ are the mean, rotational and entropy part of $\KLM{.}{q}$}
        	  
        	  \vspace{0.1cm}
              \State $\Sigma \leftarrow \eta_g \Sigma + (1-\eta_g) \Sigma_q$\label{alg:line:cov}
           \EndIf \vspace{0.2cm}
           \If{$\EEc{s}\left[\KLM{\mathcal{N}(M\psi(s), \Sigma)}{q(.|s)}\right] > \epsilon$}\label{alg:line:check2}
              \State $a \leftarrow \dfrac{1}{2}\EEc{s\in\mathcal{T}}\lVert M\psi(s)-M_q\psi(s)\rVert^2_{\Sigma_q^{-1}}$ \vspace{0.2cm}
        	  \State $b \leftarrow \dfrac{1}{2}\EEc{s\in\mathcal{T}}[(M\psi(s) - M_q\psi(s))^T\Sigma_q^{-1}(M_q\psi(s)-M_q\psi_q(s))]$ \vspace{0.2cm}
              \State $c \leftarrow m_q(M_q) + r_q(\Sigma) + e_q(\Sigma) - \epsilon$  \vspace{0.1cm}
        	  \State $\eta_m \leftarrow \dfrac{-b + \sqrt{b^2 - ac}}{a}$ \vspace{0.1cm}
              \State $M = \eta_m M + (1 - \eta_m) M_q$\vspace{0.1cm}\label{alg:line:mean}
           \EndIf
           \State $\pi^c_{\theta}(.|s) \leftarrow \mathcal{N}(M\psi(s), \Sigma)$ \vspace{0.1cm}
           \State \Return $\pi^c_{\theta}(.|s)$\label{alg:line:return}
       \EndProcedure
    \end{algorithmic}
\end{algorithm}

In this paper, we will make use of the \gls{papi} techniques proposed in \cite{Akrour19}. 
These projections can be used to transform a constrained optimization problem, such as \gls{rl} under entropic constraints, to an unconstrained one. The core idea of the \gls{papi} approach is to find a projection $g$ that maps the parametric policy space to a subspace thereof satisfying the constraints. The constrained maximization of the policy update objective $f$ is then solved by an unconstrained maximization of $f \circ g$. The mapping $g$ is selected to be differentiable, and thus, standard gradient descend techniques can be applied to the resulting optimization problem. This technique is similar to the Projected Gradient method~\cite{bertsekas1999}, that projects back to the acceptable region after each gradient step, while in \gls{papi} the projection is incorporated into the optimization.

In our setting we make use of Gaussian policies, and update them in a policy iteration setting under a KL-divergence and entropy constraints, as introduced in Sec.~\ref{sec:diffopt}. As such, we will make use of Alg.~\ref{alg:linproj}, which is a special case of the Alg~2 in~\cite{Akrour19} that projects the parameters of any linear-Gaussian---i.e. a Gaussian distribution for which the mean is a linear function of some state features $\psi$--- of shape $\pi(a|s) = \mathcal{N}(a | M\psi(s), \Sigma)$, to a linear-Gaussian that complies with a KL-divergence and entropy constraints. Since this projection is differentiable, the composition of the objective $f$ of the policy update and the projection can then be optimized using gradient ascent in an unconstrained way. In more details, note that the projected policy returned by Alg.~\ref{alg:linproj} in Line~\ref{alg:line:return} is a mixture of the original policy parameters $\theta$ and the policy parameters of the previous policy $q$. The projection proceeds by first ensuring that the entropy constraint is respect for the input covariance matrix~(Line~\ref{alg:line:entropy}). Then it checks whether the KL-divergence constraint is violated~(Line~\ref{alg:line:check1} and \ref{alg:line:check2}) before interpolating the parameters of the input linear-Gaussian policy with the parameters of the previous policy~(Line ~\ref{alg:line:cov} and \ref{alg:line:mean}). The interpolation parameters $\eta_g$ and $\eta_m$ are computed in closed form from upper bounds of the constraint. Full details on the linear-Gaussian projection and proofs that Alg.~\ref{alg:linproj} returns policy parameters that comply with the KL-divergence and entropy constraints can be found in~\cite{Akrour19}.
    	  
    	
	

\section{Mixture of interpretable experts policy}
\label{sec:pol}
We describe more formally in this section the policy structure and the associated learning algorithm. The policy samples an action by comparing the current state to a list of fixed size K of cluster centers ${\cal C} = \{s_0,\dots, s_{K-1}\}$. It can then, for instance, select the action associated with the closest cluster. This would yield a discrete optimization problem. However, learning a satisfactory policy in this setting is a challenging problem. We relax our model, by considering instead a smoothed version of the previous problem with fuzzy memberships~\cite{zadeh1965fuzzy} to each cluster, akin to a mixture of expert approach \cite{Masoudnia14}. 

Formally, the policy at state $s$ is a Gaussian distribution $\pi(a|s) = \mathcal{N}(a | \mu(s), \Sigma)$ with mean given by $\mu(s) = M\psi(s)$, where $M$ is a $K\times \text{dim}(\mathcal{A})$ matrix containing the associated action to each cluster or expert, and $\psi(s)$ is a feature vector of $s$ describing distance to each prototypical states. Specifically
\begin{align}
    \psi(s) &\delequal \dfrac{w(s)}{\norm{w(s)}_1 +1},\label{eq:psimean}\\
    w(s) &\delequal c\odot \varphi(s),\label{eq:wmean}\\
    \varphi_i(s)&\delequal\exp(-\tau\norm{s - s_i}_2^2),\label{eq:disttoproto}
\end{align}
with $w(s)$ the vector of unnormalized cluster memberships, $\norm{.}_1$ the $\ell_1$ norm, $c$ is the cluster weights parameter vector, $\varphi(s)$ is the vector of distances to each cluster center, and the operator $\odot$ represents the Hadamard, i.e. element-wise, product.
For each cluster center $s_i$, the distance is computed as in Eq.~\eqref{eq:disttoproto},
where $\norm{.}_2$ is the Euclidian norm and $\tau$ is a fixed temperature parameter.

A final component of the policy is the default action. When a state is far from all cluster centers $\{s_i\}_{i=1}^K$, the policy becomes too sensitive to small changes in the state variables, as all $\varphi_i(s)$ and hence $w_i(s)$ in the denominator of Eq.~\eqref{eq:psimean}, are close to zero. This sensitivity introduces both numerical problems and hinders interpretability. We thus introduce a default action, with unnormalized membership fixed to one, independently of the input state. For simplicity we will assume that the value of the default action is fixed to the null vector. This leaves $M$ unchanged and simply requires the additional $1$ shown in the denominator of Eq.~\eqref{eq:wmean}. The default action has an added benefit in terms of interpretability, as one is able to know when the policy is presented with an unfamiliar state, in which case the sum of the normalized membership features $\sum_i\psi_i(s)=\sum_i\frac{w_i(s)}{\norm{w(s)}_1+1}$ will be close to zero. In contrast, without the default action, the sum $\sum_i\frac{w_i(s)}{\norm{w(s)}_1}$ will always be equal to one.

The policy structure is simple and the main challenge that it presents is to find an appropriate set of cluster centers $\cal C$ during learning. As we strive for human-readable policies, the difficulty will be to only select elements of $\cal C$ from trajectory data. This operation is non-differentiable and requires mixing discrete and continuous optimization. This difficulty is exacerbated by the necessity of keeping $K$ small for similar reasons of interpretability.

%

\section{Learning the mixture of interpretable experts policy}
\label{sec:opt}
\begin{algorithm}[t]
	\begin{algorithmic}[1]
    	\State \textbf{Input:} environment $\mathcal{M}$, Initial value function $V$ 
    	\For{$\text{it}\leftarrow 0$ \textbf{to} $N$}
    	    \State $\mathcal{T} \leftarrow$ \Call{generateTrajectories}{$\mathcal{M}, q$} \label{alg:initbegin}
    	    \State $A \leftarrow$ \Call{computeGAE}{$q, V$}
    	    \State $V \leftarrow$ \Call{updateValueFunction}{$V, A$}
    	    
    	    \If{$\text{it}=0$}
    	        \State $q \leftarrow$ \Call{addClusters}{$q, \mathcal{T}, A)$} 
    	    \EndIf \label{alg:initend}
    	    
    	    \If{$\text{it} \bmod 2$} \label{alg:upbegin}
    	        \State $\pi \leftarrow$ \Call{updateFullPolicy}{$q, \mathcal{T}, A$}
    	   \Else
    	        \State $\pi \leftarrow$ \Call{swapClusters}{$q, \mathcal{T}, A$}\label{alg:swap_line}
    	        \State $\pi \leftarrow$ \Call{compressPolicy}{$\pi,  q, \mathcal{T}, A$}
    	        \If{$\pi \neq q$}
    	            \State $\pi \leftarrow$ \Call{updateMeanAndCov}{$\pi, q, \mathcal{T}, A$}
    	        \Else
    	            \State $\pi \leftarrow$ \Call{updateFullPolicy}{$q, \mathcal{T}, A$}\label{alg:fullupnochange}
    	        \EndIf
    	   \EndIf
    	   \State $q \leftarrow \pi$ \label{alg:upend}
        \EndFor
    	
    	\State \Return $\pi$ 
	\end{algorithmic}
	\caption{Learning Mixture of Interpretable Experts}
	
	\label{alg:interpretable}
\end{algorithm}

Our algorithm is presented in Alg.~\ref{alg:interpretable}, and is based on the \gls{api} scheme~\cite{Bertsekas11,Scherrer14}. Line~\ref{alg:initbegin} to \ref{alg:initend} are the initialization and policy evaluation steps while Line~\ref{alg:upbegin} to \ref{alg:upend} are about the policy update. 

\subsection{Initialization and policy evaluation}
\label{sec:initnpe}
The policy is initialized with the first state encountered as an initial cluster. All cluster weights vector are initialized to zero, excluding the weight for the first cluster, which  is initialized to 1. The cluster action means are initialized to zero. 
After generation of the first iteration's trajectories and after the advantage function is estimated, the other cluster centers are initialized by selecting the $K-1$  clusters and cluster actions from the state-action couples with the highest advantage value in the collected dataset. This operation does not change the policy distribution as the cluster weights for the uninitialized clusters are set to zero.

Regarding policy evaluation, the sampling and the \gls{gae} are performed in Alg.~\ref{alg:interpretable} by the \texttt{generateTrajectories} and the \texttt{computeGAE} procedures respectively. The initial cluster center selection is performed by the \texttt{addClusters} procedure.
As done in other standard \gls{api} algorithms~\cite{Schulman15,Schulman17}, we generate trajectories rollouts from the environment before estimating the advantage function using the \gls{gae} algorithm~\cite{schulman16}. What remains, given the advantage value estimated for every state-action pair in the dataset is to update the policy.

\subsection{Policy update}
The policy update can be divided into two optimization categories, a differentiable optimization problem where we optimize the policy parameters that are the cluster action matrix $M$, the cluster weights $c$, and the policy variance $\Sigma$, and a discrete optimization problem where we update the cluster center lists $\cal C$. In the following, we  describe these two optimization problems in details.

\subsubsection{Discrete optimization}
\label{sec:discopt}
In the discrete optimization phase, we seek to update the cluster list $\cal C$, which is the basis for the computation of the cluster distances $\varphi$ and the policy mean $\mu(s)$. To retain the interpretable nature of our policy, we enforce the cluster list to contain only states generated by the environment. To illustrate the importance of this constraint, consider an autonomous driving car policy that has an image as input. Allowing the cluster centers to be optimized by gradient descent will result in cluster centers exiting the manifold of human understandable images. In contrast, enforcing the cluster list to only contain images encountered by the autonomous car will facilitate the analysis of the policy by a human and show which prototypical real-world situations make the autonomous car decide on a particular action.

However, as the set of possible states is a discrete set, optimizing the cluster list becomes more challenging and the policy cannot be trained 'end-to-end' by gradient descent. As such we separate the optimization of $\cal C$ from the optimization of $M$, $c$ and $\Sigma$. To optimize $\cal C$ we need to answer two questions, which objective to optimize and with which optimization algorithm. 

\paragraph{Optimization objectives}
We have evaluated several such objectives, including the maximization of the advantage value, a typical choice in policy update. However, we have obtained better results with objectives that favor the spread of the cluster centers and their coverage of the state space, while performing the maximization of the advantage during the differentiable optimization phase.

To maximize coverage we have first considered to maximize $\EEc{s\sim q}\sum_i\frac{w_i(s)}{\norm{w(s)}_1+1}$, i.e. to have cluster centers spread in such a way that the default action is executed as infrequently as possible. However, this has resulted in having several clusters clamped close to each other, possibly around states frequently visited by $q$, to reduce the influence of the default action. Instead, a better solution would be to keep one cluster in the same area and increase its cluster weight in the differentiable optimization phase. 

The optimization criterion that ended-up providing the best results fulfills the following two criteria: it does not depend on parameters optimizer in the differentibale optimization phase and it penalizes clusters being close to each other. This criterion is given by
\begin{align}
\EEc{s\sim q}\left[\varphi_{t_1}(s)-\frac{1}{N}\sum_{i=1}^N\varphi_{t_i}(s)\right],\label{eq:swapobj}
\end{align}
where $t$ is an ordering of $\varphi(s)$ in descending order such that $\varphi_{t_1}(s) = \max_i\varphi_i(s)$, $\varphi_{t_2}(s)$ has the second highest value and so on. For the hyper-parameter $N$ we have tried three values $N=2$, $N=3$ and $N=K$ and the best performance was achieved with $N=3$. This objective maximizes the margin between the highest value over $i$ of $\varphi_i(s)$ and the average value over $i$ of $\varphi_i(s)$. A high margin implies that $s$ is close to one and only one cluster center.

\paragraph{Optimization technique} 
Regarding the discrete optimization of the objective, the high number of possible candidates, that is the set of all states encountered so far or even just the states generated by the current policy, prevents an exhaustive search of all possible combinations, that can be exponential in the number of candidate states. Another aspect complicating the optimization process is the \gls{kl} constraint later introduced in Sec.~\ref{sec:diffopt} that precludes the use of submodular optimization techniques because it does not define a matroid structure \cite{rajaraman2018submodular}. To tackle this challenging optimization problem we resort to approximate techniques by using random search and systematically check for the satisfaction of the constraint.

The \texttt{swapClusters} routine (in Line~\ref{alg:swap_line} of Alg.~\ref{alg:interpretable}) will perform a random search in the cross set of states sampled by $q$ and the states in the current cluster list $\cal C$. The routine is composed of a two-step randomization scheme. In the first step, a prioritized sampling using a heuristic function $h_1$ is used to sample a set of $k$ cluster centers from $\cal C$ to be swapped with $k$ state candidates sampled from the generated trajectories $\cal T$, using a similar sampling algorithm but with a heuristic $h_2$. 

Both steps are based on the polynomial randomization scheme presented in~\cite{bresina1996heuristic}. 
The heuristic $h_1$ to select the cluster center to be replaced and the heuristic $h_2$ to sample possible candidates are given by
\begin{align}
    h_1(s_i) &\delequal \EEc{s\sim q}\left[w_i(s)\norm{w(s)}_1^{-1}\right],\\
    h_2(s) &\delequal \norm{\varphi(s)}_1.
\end{align}

The heuristic $h_1$ favors clusters with low activation~(candidates are ranked in descending order, lower heuristic value is better). We take into account the cluster weights since $h_1$ is also about selecting clusters that are not likely to violate the KL-divergence constraint if swapped, and the KL-divergence greatly depends on the cluster weight. On the other side, $h_2$ favors states that are far from the current cluster centers.

Using this randomization scheme we generate a set of $n$ candidates. For every candidate, we check that the obtained policy does not violate the KL-divergence constraint and pick the one with the highest objective in Eq.~\eqref{eq:swapobj}. To perform the maximum number of swaps, we repeat the sampling multiple times, starting from $K$ total swaps i.e., swapping all clusters, and halving the number of swaps if the randomization cannot improve w.r.t. the objective in Eq.~\eqref{eq:swapobj}.

\paragraph{Post processing}
Despite the objective in Eq.~\eqref{eq:swapobj} penalizing overlap of clusters, it can still happen that the clusters are too close to each other which hinders the expressiveness of the policy. To overcome this problem we add a routine \texttt{compressPolicy} that is executed after the swapping routine. The compression takes a simple form, where we try to delete every cluster by setting its cluster weight to 0, and keeping the resulting policy if the KL-divergence constraint is not violated.

\subsubsection{Differentiable optimization}
\label{sec:diffopt}
The policy parameters---cluster weights, cluster actions, and covariance---are updated by solving the following constrained optimization problem

\begin{alignat}{3}
& \argmax_\pi && L(\pi),\label{eq:objapi}\\
& \text{subject to}\ \ \ \ \ 
&& \EEc{s \sim q}\left[\KLM{\pi(.|s)}{q(.|s)}\right] &&\leq \epsilon,\label{eq:KLapi}\\
& &&\EEc{s\sim q} \left[\HH(\pi(.|s))\right]   &&\geq \beta,\label{eq:entropyapi} 
\end{alignat}

where $q$ is the data generating policy. The optimization problem is similar to the one presented in~\cite{Schulman15}. The objective of the problem $L(\pi) = \EEc{s,a \sim q}\left[\frac{\pi(a|s)}{q(a|s)} A_q(s,a)\right]$ is to maximize the expected advantage function. The constraints are a \gls{kl}-divergence constraint between successive policies---akin to a step-size---and an entropy constraint---to sustain exploration. The entropy constraint is especially important in our case since the policy expressiveness is initially quite limited. Indeed, as described in Sec.~\ref{sec:initnpe}, the cluster center list $\mathcal{C}$ initially only contains the first encountered state as an active cluster---i.e. with non-zero cluster weights. Policy improvements on the objective will initially be quite modest compared to a more intricate model and the entropy constraint helps in sustaining exploration while a more representative cluster center list $\cal C$ is learned.

\paragraph{Projection for mixture of expert policy} To tackle the constrained optimization problem we extend the projections for \gls{papi} techniques to our specific policy structure. Specifically, we aim to extend the linear-Gaussian projection discussed in the preliminaries.

The projection described in Alg.~\ref{alg:linproj} works by linearly interpolating the parameters of the input policy, $M$ and $\Sigma$, with the parameters of the data generating policy $q$. To be able to optimize our policy with similar techniques, we need to provide a projection of all the policy parameters, including the newly introduced cluster weights $c$. Since the cluster weights in our policy only affect the mean of the Gaussian, the entropy constraint in Eq.~\ref{eq:entropyapi} will not be affected by a change in cluster weight and the same projection as in \cite{Akrour19} can be used for $\Sigma$ to ensure that the entropy constraint is satisfied. As for the KL, since $\pi(.|s)$ and $q(.|s)$ are Gaussians, their KL can be expressed in closed form and is given by 
\begin{align}
\KLM{\pi(.|a)}{q(.|a)} = \frac{1}{2}m(\mu(s)) + \frac{1}{2}r(\Sigma) + \frac{1}{2}e(\Sigma),
\end{align}
where $m(\mu(s)) = ||\mu(s)-\mu_q(s)||^2_{\Sigma_q^{-1}} = (\mu(s)-\mu_q(s))^T \Sigma_q^{-1}(\mu(s)-\mu_q(s))$ is the change in mean, $r(\Sigma) =\text{tr}(\Sigma_q^{-1}\Sigma) - d$ is the rotation of the covariance and $e(\Sigma)=\log\frac{|\Sigma_q|}{|\Sigma|}$ is the change in entropy. In the KL, only $m$ depends on $s$. For clarity of notations, we drop the dependence on $s$ for now, and further expand $m$ into
\begin{align}
m(\mu) = \left\lVert M\dfrac{w}{\lVert w\rVert_1} - M_q\dfrac{w_q}{\lVert w_q\rVert_1} \right\rVert_{\Sigma_q^-1}^2,
\end{align}
where we have also dropped for clarity, and without loss of generality, the default action contribution, that can be thought of as integrated to $w$. 

Now, given the cluster weights $c$ and cluster actions $M$ of an arbitrary policy that violates the constraint in Inq.~\eqref{eq:KLapi}, we want to find a 'projected' policy $\pi'$ that has new cluster weights and cluster actions of the form $c_\eta = \eta c + (1-\eta)c_q$ and $M_\nu = \nu M + (1-\nu)M_q$ such that $\EEc{s \sim q}\left[\KLM{\pi'(.|s)}{q(.|s)}\right] \leq \epsilon$. To do so, we will write the constraint in Inq.~\eqref{eq:KLapi} in terms of the interpolation parameters $\nu$ and $\eta$, and solve for these parameter $\EEc{s \sim q}\left[\KLM{\pi'(.|s)}{q(.|s)}\right] = \epsilon$. We know that such a solution exists since the KL-divergence is zero for $\eta = \nu = 0$, is $>\epsilon$ for $\eta = \nu = 1$ and is continuous in $\eta$ and $\nu$. Unfortunately, efficiently solving this equation is not possible since no closed form exists. However, it is important that the projection is efficient to compute since it is called several times during the policy update. What we propose instead is to derive an upper bound $u(\nu, \eta)$ of $\EEc{s \sim q}\left[\KLM{\pi'(.|s)}{q(.|s)}\right]$ that has a simpler form in $\eta$ and $\nu$ and solve for these parameters the equation $u(\nu, \eta) = \epsilon$ in closed form and in an efficient manner.

\begin{figure*}[t]
	\begin{subfigure}[t]{0.24\textwidth}
		\includegraphics[width=\textwidth]{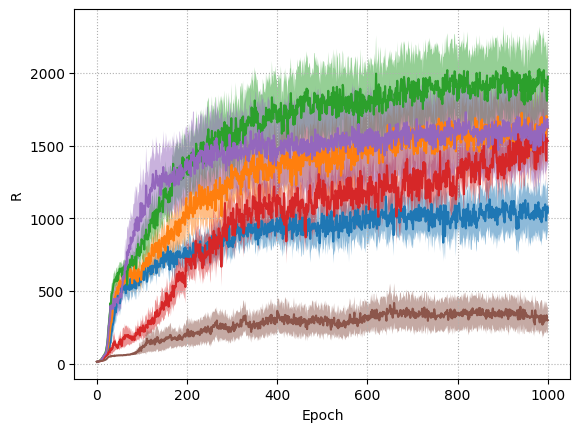}
		\caption{HopperBulletEnv-v0}
	\end{subfigure}
	\hfill
	\begin{subfigure}[t]{0.24\textwidth}
		\includegraphics[width=\textwidth]{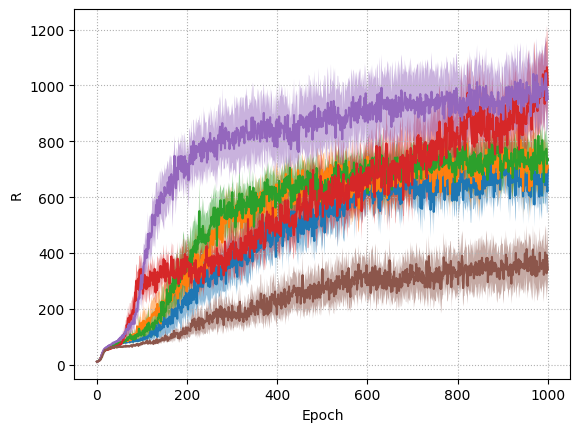}
		\caption{Walker2DBulletEnv-v0}
	\end{subfigure}
	\begin{subfigure}[t]{0.24\textwidth}
		\includegraphics[width=\textwidth]{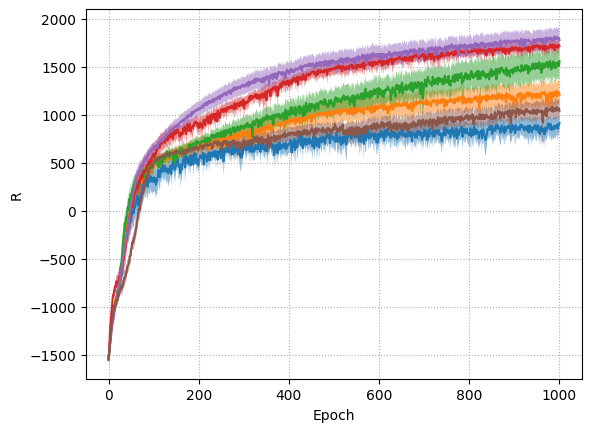}
		\caption{HalfCheetahBulletEnv-v0}
	\end{subfigure}
	\hfill
	\begin{subfigure}[t]{0.24\textwidth}
		\includegraphics[width=\textwidth]{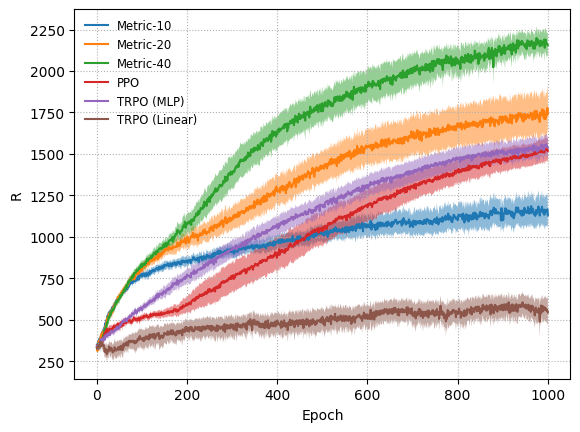}
		\caption{AntBulletEnv-v0}
	\end{subfigure}
	\caption{Performance of our algorithm on four \texttt{pybullet} locomotion tasks with 10, 20 and 40 clusters. Our algorithm is compared to PPO, and TRPO with neural network and linear policy. All plots averaged over 25 runs, showing mean and 95\% confidence interval.}
	\label{fig:perf}
\end{figure*}
\begin{figure*}
	\captionsetup{format=hang}
	\centering
	\begin{subfigure}[t]{0.24\textwidth}
		\includegraphics[width=\textwidth]{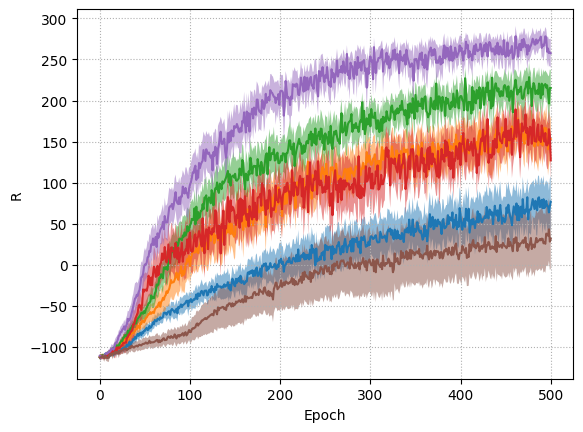}
		\caption{BipedalWalker-v2}
	\end{subfigure}
	\hfill
	\begin{subfigure}[t]{0.24\textwidth}
		\includegraphics[width=\textwidth]{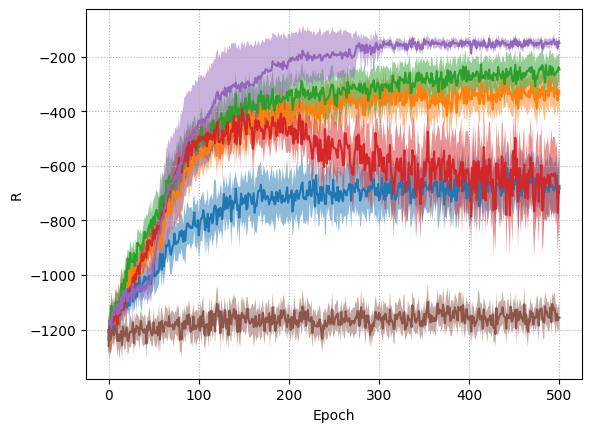}
		\caption{Pendulum-v0}
	\end{subfigure}
	\hfill
	\begin{subfigure}[t]{0.24\textwidth}
		\includegraphics[width=\textwidth]{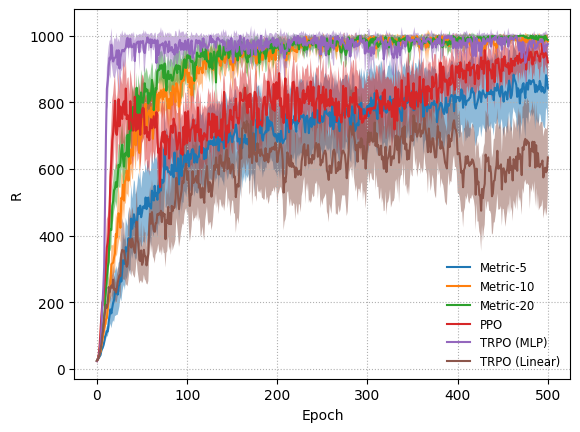}
		\caption{InvertedPendulum\allowbreak BulletEnv-v0}
	\end{subfigure}
	\caption{Performance of the interpetable algorithm on three additional Openai-Gym environments. Our algorithm is run with 5, 10 and 20 clusters, and compared to TRPO and PPO with a neural policy and TRPO with a linear policy. Plots show the average reward and $95\%$ confidence interval, out of 25 independent runs.}
	\label{fig:smallres}
\end{figure*}

Since only $m$ in the KL-divergence depends on the policy mean, we let $m(\nu, \eta)$ be the change in mean for $\pi'$, i.e.
\begin{align}
    m(\nu, \eta) = & \left\lVert M_\nu\dfrac{w_\eta}{\lVert w_\eta\rVert_1} - M_q\dfrac{w_q}{\lVert w_q\rVert_1} \right\rVert_{\Sigma_q^-1}^2,\label{eq:mvn}
\end{align}
where again we have dropped for now the dependency on $s$. Note that in Eq.~\eqref{eq:mvn}, we are considering the interpolation parameter $\eta$ in terms of the unnormalized cluster memberships $w$ instead of the cluster weight $c$. This is perfectly equivalent due to the linear dependency between $c$ and $w$.

By expanding the square, $m(\nu, \eta)$ can be decomposed into
{\small
\begin{flalign}
    m(\nu, \eta) = &\nu^2\left\lVert \left( M - M_q \right)\dfrac{w_\eta}{\lVert w_\eta\rVert_1}\right\rVert_{\Sigma_q^-1}^2 + 2 \nu \nonumber\\
    &\left(\left( M - M_q \right)\dfrac{w_\eta}{\lVert w_\eta\rVert_1} \right)^T\Sigma_q^{-1}\left( M_q\left(\dfrac{w_\eta}{\lVert w_\eta\rVert_1} - \dfrac{w_q}{\lVert w_q\rVert_1}\right) \right)\nonumber\\
    &+ \left\lVert M_q\left(\dfrac{w_\eta}{\lVert w_\eta\rVert_1} - \dfrac{w_q}{\lVert w_q\rVert_1}\right) \right\rVert_{\Sigma_q^{-1}}^2,
\end{flalign}
}
where the first term quantifies the change of cluster actions in $M$, the last term the change in cluster weights $c$---and hence $w$---and the second term is a cross term. The last term is equal to $m(0, \eta)$ and can be further decomposed into 
\begin{align}
    m(0, \eta) = \dfrac{\eta^2\lVert w \rVert^2_1}{\lVert \eta w +(1-\eta)w_q \rVert_1^2}m(0,1).
\end{align}
$m(0, \eta) = \epsilon$ does not admit an easy solution but we derive the following two upper bounds for it
\begin{align}
\label{eq:max}
m(0, \eta) &\leq \eta^2\max\left(\dfrac{\lVert w \rVert^2_1}{\lVert w_q \rVert_1^2}, 1\right)m(0,1),\\
\label{eq:conv}
m(0, \eta) &\leq \eta \frac{\norm{w}_1^2}{2\norm{w_q}_1\norm{w}_1-\norm{w_q}_1^2} m(0,1).
\end{align}
Proofs for both upper bounds are given in the appendix. Inq.~\eqref{eq:max} always holds while Inq.~\eqref{eq:conv} only holds if the denominator is positive. Both bounds have now an easy solution $\eta$, as they are quadratic or linear functions of $\eta$. We use both the bound in Inq.~\eqref{eq:max}, which is tighter when $\norm{w}_1 \approx \norm{w_q}_1$, and the bound in Inq.~\eqref{eq:conv}, which is tighter when $\norm{w}_1 \gg \norm{w_q}_1$ by simply taking the one that offers the largest $\eta$, i.e. for which the KL-divergence of $\pi'$ will be the closest to $\epsilon$.

The upper bounds of $m(0, \eta)$ are derived for a single $s$ but the KL-divergence constraint is in expectation of states generated by $q$. Letting $m_s$ be the mean shift of the KL-divergence component at state $s$, bounding the expected KL-divergence w.r.t. states of $q$ will require to bound $\EEc{s \sim q}\left[m_s(0,\eta)\right]$. However, the expectation still yields upper bounds easy to solve for, since $\eta^2$ and $\eta$ can simply be taken out of the expectation and the remaining quantities can be evaluated from trajectories.

Being able to find an $\eta$ s.t. $\EEc{s \sim q}\left[m_s(0,\eta)\right] \leq \epsilon$ gives us now a path to find a projected policy that complies with the KL-divergence constraint. Given arbitrary cluster weights $c$ and cluster actions $M$ encountered during gradient ascent in the policy update, we define the projection as follow: first fix $M$ to $M_q$ and find an $\eta$ solution of $\EEc{s \sim q}\left[m_s(0,\eta)\right]\leq \epsilon$. Given this $\eta$, fix the cluster weight to $c_\eta$ and compute the new state features resulting from the change in cluster weights $\psi(s) = \dfrac{w_\eta(s)}{\norm{w_\eta(s)}_1+1}$. Then use Alg.~\ref{alg:linproj} to project $M$ and $\Sigma$, the remaining parameters of the linear-Gaussian policy. Indeed, by finding $\eta$ such that $\EEc{s \sim q}\left[m_s(0,\eta)\right]\leq \epsilon$, we have updated the cluster weights, and hence the features of the policy, while ensuring that the requirements of Alg.~\ref{alg:linproj}---namely that the KL-divergence constraint is not violated upon changing the features---are satisfied. As such, the projection for the remaining parameters $M$ and $\Sigma$ of the policy follow as in Alg.~\ref{alg:linproj}.  

\paragraph{Optimization with projections.}
The optimization algorithm is presented in Alg.~\ref{alg:interpretable}.
Due to the presence of a concurrent discrete optimization process, the differentiable optimization has to be performed carefully, in order to ensure that the constraints are fulfilled.

When the cluster list $\cal C$ is modified, we cannot apply the projection for the cluster weights. Indeed, even if Inq.~\eqref{eq:max} and Inq.~\eqref{eq:conv} still hold and we can find an $\eta$ to bring $m$ under $\epsilon$, we cannot translate the interpolation in $w$ to an interpolation in $c$ anymore since the feature function $\varphi$ changed. As such, we resort in this case to only updating $M$ and $\Sigma$ of the policy in  \texttt{updateMeanAndCov}. To ensure that the cluster weights are learned, we only update the cluster list every second iteration and perform a full update of all differential parameters of the policy otherwise. If the discrete optimization does not modify the cluster list, as shown in Line~\ref{alg:fullupnochange} of the algorithm, a complete update of the policy parameters---including the cluster weights---is performed instead in \texttt{updateFullPolicy}. 

Because of the change to the cluster list $\cal C$, simpler and more common approaches such as TRPO \cite{Schulman15}, that perform policy update under a KL-divergence constraint, cannot be used in our case. Indeed, the theoretical framework of TRPO requires that the KL-divergence constraint is approximated around the data generating policy $q$ whereas we want to update the parameters of a policy with a different cluster list. This policy is not $q$ anymore even when all differentiable parameters are the same. Even though the complexity of the proposed policy update might not be desirable, its payoff is an algorithm that is competitive with TRPO on many \gls{rl} tasks even when a simpler policy is being used.

\section{Experimental Evaluation}
\begin{figure*}[t]
	\centering
	\begin{subfigure}[t]{0.2\textwidth}
		\includegraphics[width=\textwidth]{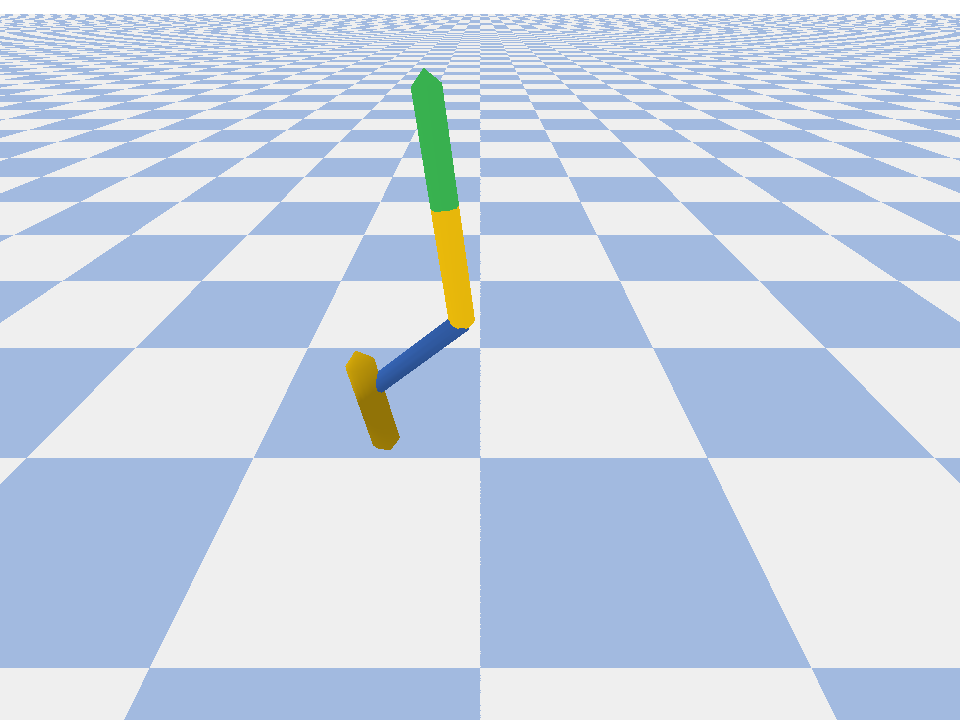}
		\caption{Cluster 0}
	\end{subfigure}
	\hfill
	\begin{subfigure}[t]{0.2\textwidth}
		\includegraphics[width=\textwidth]{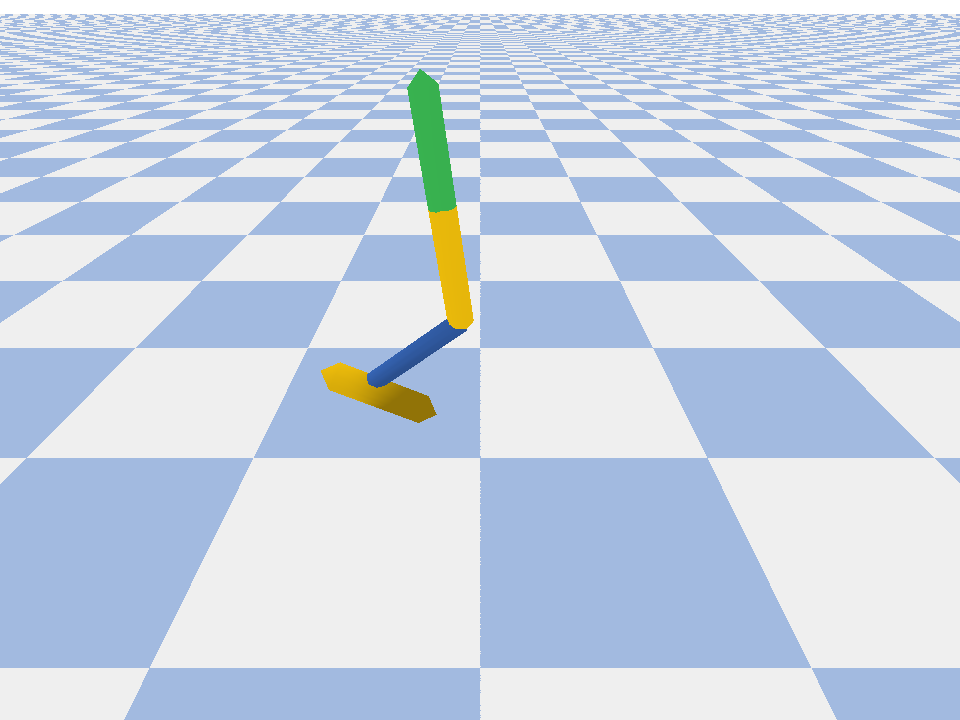}
		\caption{Cluster 9}
	\end{subfigure}
	\hfill
	\begin{subfigure}[t]{0.2\textwidth}
		\includegraphics[width=\textwidth]{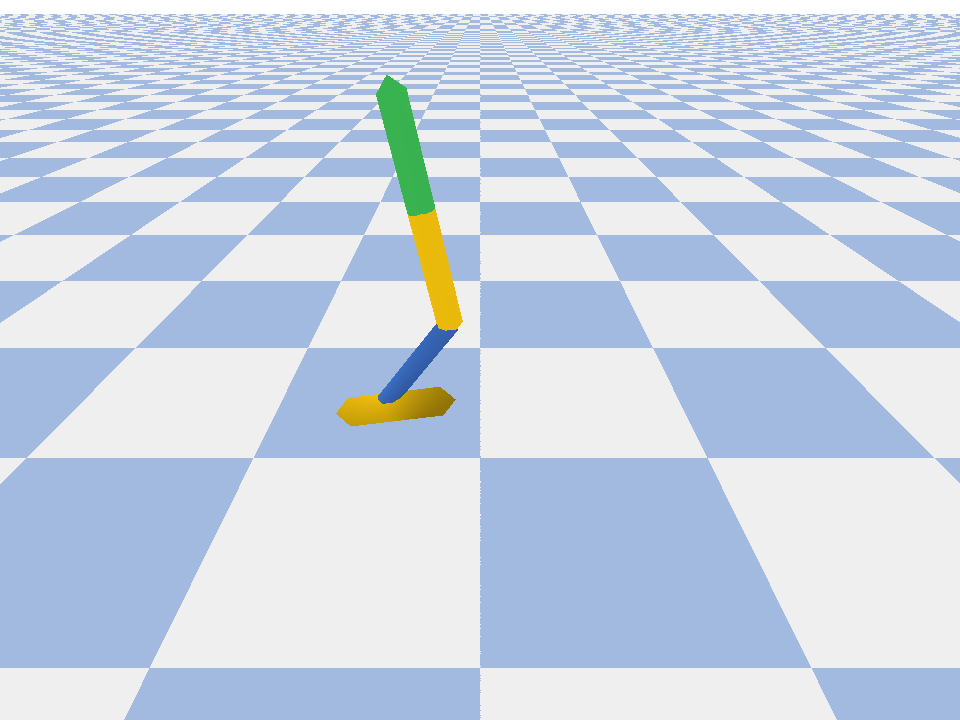}
		\caption{Cluster 4}
	\end{subfigure}
	\begin{subfigure}[t]{\textwidth}
	    \centering
		\includegraphics[width=.8\textwidth]{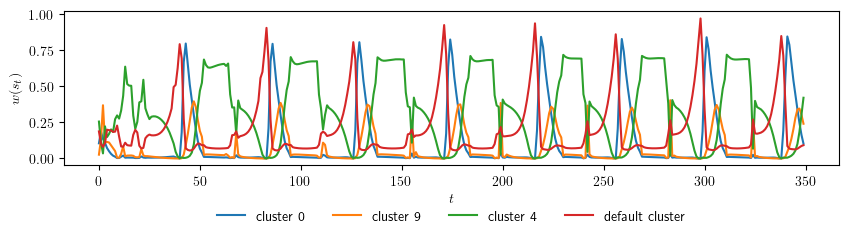}
		\caption{Cluster activation}
		\label{fig:sub:hopact}
	\end{subfigure}
	
	\caption{Three clusters out of ten and their associated activation on the \texttt{HopperBulletEnv-v0} environment learned by our algorithm.}
	\label{fig:hopact}
\end{figure*}

\begin{figure*}[t]
	\centering
	\begin{subfigure}[t]{\sizechetah\textwidth}
		\includegraphics[width=\textwidth]{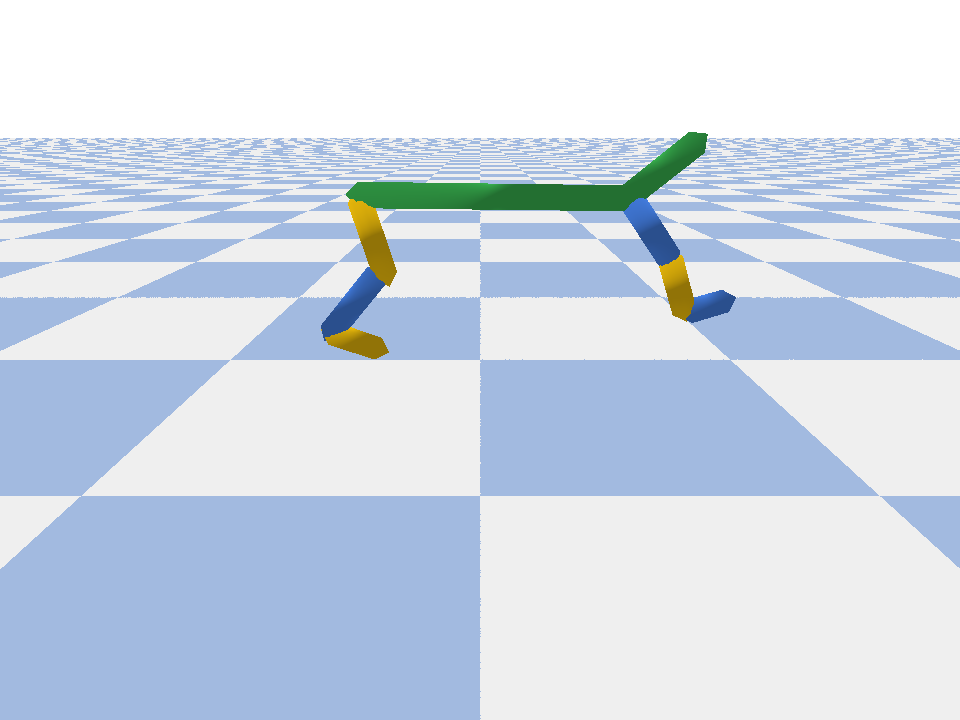}
		\caption{Cluster 4}
	\end{subfigure}
	\hfill
	\begin{subfigure}[t]{\sizechetah\textwidth}
		\includegraphics[width=\textwidth]{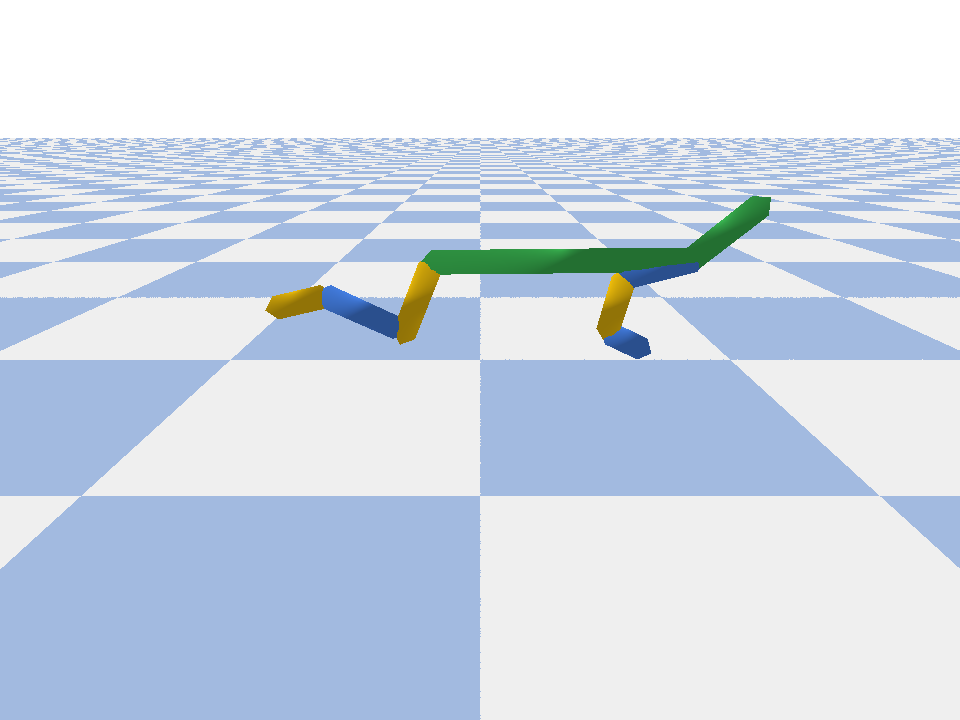}
		\caption{Cluster 2}
	\end{subfigure}
	\hfill
	\begin{subfigure}[t]{\sizechetah\textwidth}
		\includegraphics[width=\textwidth]{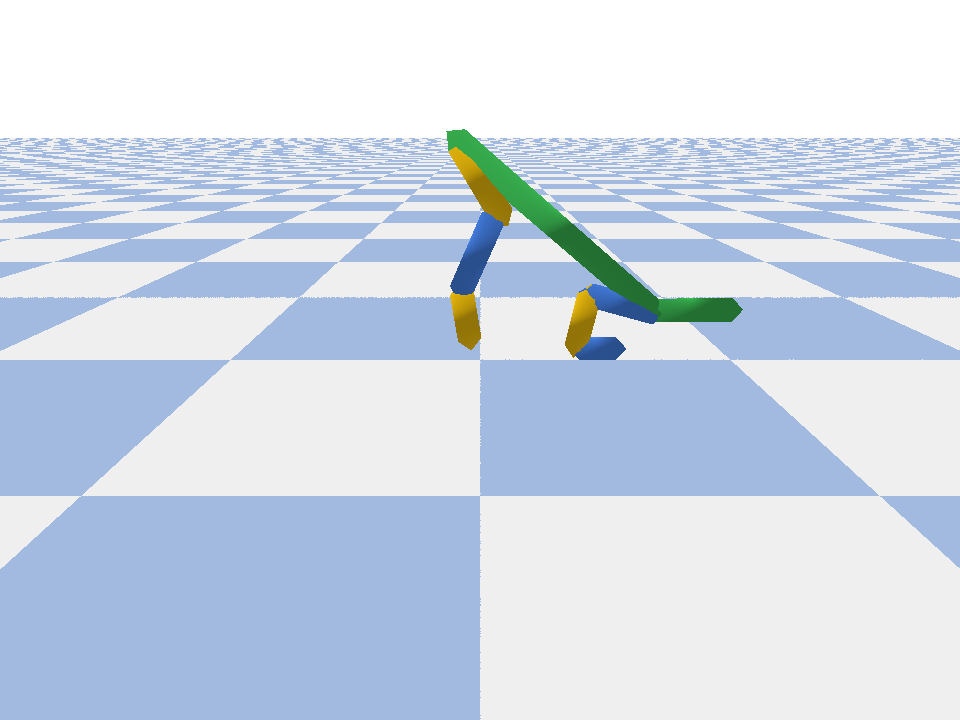}
		\caption{Cluster 9}
	\end{subfigure}
	\hfill
	\begin{subfigure}[t]{\sizechetah\textwidth}
		\includegraphics[width=\textwidth]{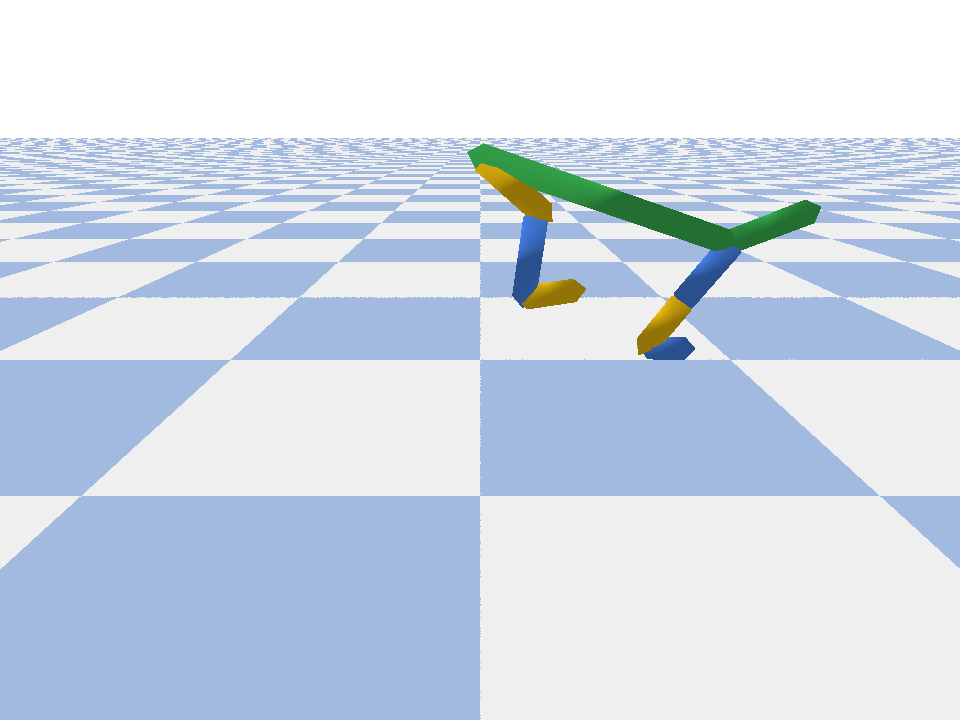}
		\caption{Cluster 0}
	\end{subfigure}
	\begin{subfigure}[t]{\textwidth}
	    \centering
		\includegraphics[width=.8\textwidth]{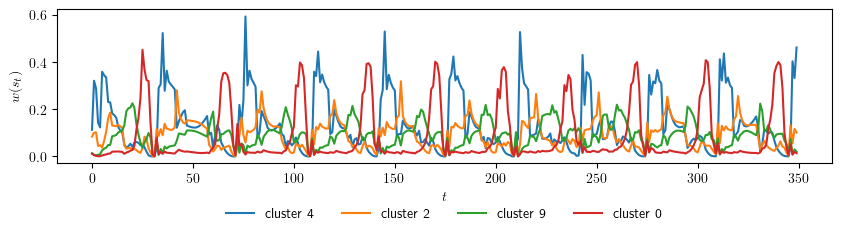}
		\caption{Cluster activation}
		\label{fig:sub:cheetact}
	\end{subfigure}
	\caption{Four clusters out of ten and their associated activation on the \texttt{HalfCheetahBulletEnv-v0} environment learned by our algorithm.}
	\label{fig:cheeact}
\end{figure*}
The objectives of this section are to assess on one hand the performance of our algorithm as an \gls{rl} learner in comparison to existing methods, and on the other hand the extent with which the returned policies can be understood and are human readable. We compare our algorithm to three RL baselines on 7 standard continuous action tasks, and analyze the interpretability of the policy on the four higher dimensional locomotion tasks of \texttt{pybullet}~\cite{coumans2019}. All code is available at \url{https://github.com/akrouriad/tpami_metricrl}.

\subsection{Performance evaluation}
We compare our algorithm to TRPO \cite{Schulman15} as it implements a similar, KL-constrained, policy update and to PPO \cite{Schulman17}. Both PPO and TRPO are learning a two hidden layer policy with 64 neurons in each layer, as this is the default for these tasks. In addition to the neural network policy, we learn using TRPO a linear policy. This setting is very similar to the publication of~\cite{rajeswaran17} that demonstrated that linear-in-features---random Fourier features---policies and linear-in-state policies can perform well on \texttt{Mujoco}~\cite{todorov12} tasks. These policies where trained using natural gradient descent~\cite{kakade2000, peters08} which is close to TRPO up to a line-search operation. However, the \texttt{Mujoco} tasks are generally easier than their \texttt{pybullet} counter-parts as the models are physically heavier in the \texttt{pybullet} task. The implementation of both TRPO and PPO uses OpenAI's \texttt{baselines}~\cite{baselines}.

We will reference to Alg.~\ref{alg:interpretable} in the plots as \texttt{Metric} and experiment with three different values of the number of clusters $K$. In the easier tasks, we use $K \in \{5, 10, 20\}$ and on the higher dimensional \texttt{pybullet} locomotion tasks we use $K\in \{10, 20, 40\}$. In each experiment, the results are averaged over 25 runs. The solid line is the mean over the runs and the shaded area is the 95\% confidence interval of the mean. The plots displayed were obtained by running 5 evaluation rollouts after each policy update for each algorithm.

On smaller scale problems shown in Fig.~\ref{fig:smallres}, TRPO with a neural network performs best, albeit as the number of clusters increases to $20$ we are able to approach the performance of neural network policies. When comparing to a linear policy, our algorithm has significantly higher performance with as little as 5 clusters. On the higher dimensional locomotion tasks, shown in Fig.~\ref{fig:perf}, our algorithm is able to match the performance of TRPO and PPO despite having less parameters, but largely outperforms TRPO with a linear policy. This shows that in the space of interpretable policies, the structure proposed in this paper offers a competitive alternative to linear policies. On a task by task comparison, our algorithm performs especially well on the \texttt{AntBulletEnv-v0} environment where it consistently learns walking policies with 10 cluster centers and largely outperforms the baselines with 40 clusters. On this environment our policy only has 1488, 748 and 378 parameters for 40, 20 and 10 clusters respectively whereas the neural network has policy has 6544 and the linear policy 232 parameters. Our algorithm is also able to learn policies competitive with neural network ones on \texttt{HopperBulletEnv-v0} with only 20 clusters, while 40 are required to edge out PPO on \texttt{HalfCheetahBulletEnv-v0} and come close to TRPO. On \texttt{Walker2DBulletEnv-v0}, albeit performance is not too far off of the baselines, it is the only environment where we did not manage to learn a walking gate with up to 40 clusters and the policies only learn to stabilize the agent. We do not believe that this is directly related to the dimensionality of the problem since \texttt{AntBulletEnv-v0} is higher dimensional and  \texttt{HalfCheetahBulletEnv-v0} is only slightly lower dimensional and in both environments, walking gaits can be learned with only 10 clusters. 

Overall, by increasing the number of clusters our algorithm approaches the performance of deep RL algorithms. With the highest number of clusters we experimented with in each setting, our algorithm typically sits between PPO and TRPO. Specifically, it outperforms PPO on 5 out of 7 tasks and outperforms TRPO on only 2 of the 7 tasks. However, when considering the interpretability of the policy, we will favor policies with a smaller amount of clusters which proved to be a lot easier to inspect. For this amount of prototypical states, our algorithm is outperformed by both deep RL baselines on all tasks. In other words, to gain interpretability we have to sacrifice a bit of performance compared to neural network policies. 

\subsection{Interpretability of the policy}
In the previous subsection, we learned successful gait policies on three PyBullet environments: \texttt{HopperBulletEnv-v0}, \texttt{HalfCheetahBulletEnv-v0} and \texttt{AntBulletEnv-v0} with as little as 10 prototypical states. We investigate now whether the expert policies---i.e. the prototypical states and their associated action---are interpretable on an individual basis, and if as a whole, they explain the gait as a composition of smaller movements. 

A first observation when analyzing the learned policy is that the number of clusters that have high activation is typically much lower (less than half) than $K$. For instance, in the Hopper environment the movement relies on only three expert policies and only four are needed for the HalfCheetah. However, when setting $K$ to such lower values from the beginning, we were unable to obtain policies that successfully walk and they could at most stabilize the robot and prevent it from falling. This indicates that future improvements of the algorithm are possible but for now we resort to manually pruning redundant clusters. In the plots of this subsection, clusters are selected if they have high activation or small cross-overlap with other clusters. A detailed experimental protocol for the analysis of the learned policies and the selection of clusters is given in the appendix.

Having selected a subset of clusters, Fig.~\ref{fig:hopact} and \ref{fig:cheeact} show their respective activation during a rollout on the Hopper and HalfCheetah environments. In both cases, the cyclical nature of the movement is clearly visible from the activation of the expert policies in Fig.~\ref{fig:sub:hopact} and  \ref{fig:sub:cheetact}. To inspect the expert policies and understand the contribution of each one of them to the overall gait, we found animated images to work best. Indeed, by initializing the agent to the state of a prototype and executing the action associated to the prototype for 5 time-steps, we are able to jointly observe the prototypical state and the effect of the associated action of each expert policy. In Fig.~\ref{fig:hopact} and \ref{fig:cheeact}, we are only able to show prototypical states in stills. Animated images are provided in the supplementary for visualizing both the prototypical state and the associated action of each expert policy.

\begin{figure*}
	\begin{subfigure}[t]{0.24\textwidth}
		\includegraphics[width=\textwidth]{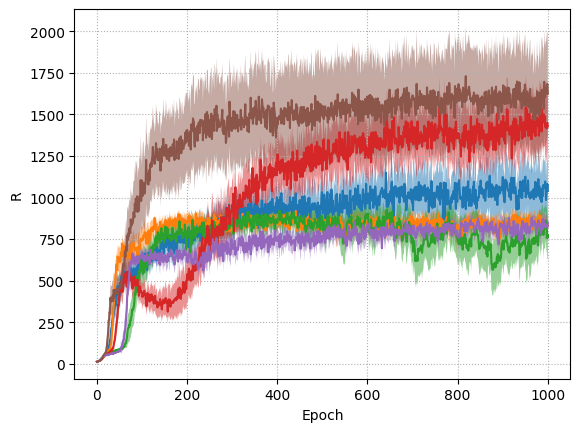}
		\caption{HopperBulletEnv-v0}
	\end{subfigure}
	\hfill
	\begin{subfigure}[t]{0.24\textwidth}
		\includegraphics[width=\textwidth]{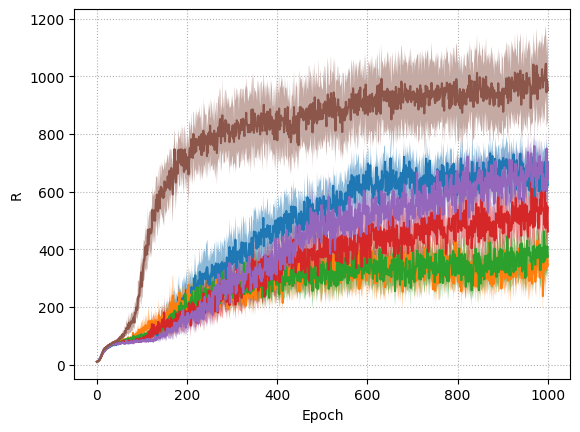}
		\caption{Walker2DBulletEnv-v0}
	\end{subfigure}
	\begin{subfigure}[t]{0.24\textwidth}
		\includegraphics[width=\textwidth]{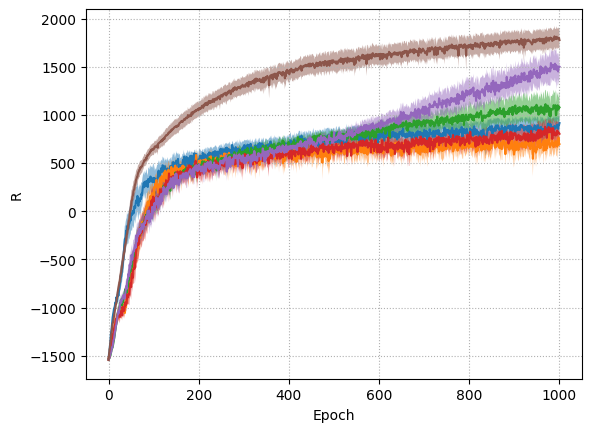}
		\caption{HalfCheetahBulletEnv-v0}
	\end{subfigure}
	\hfill
	\begin{subfigure}[t]{0.24\textwidth}
		\includegraphics[width=\textwidth]{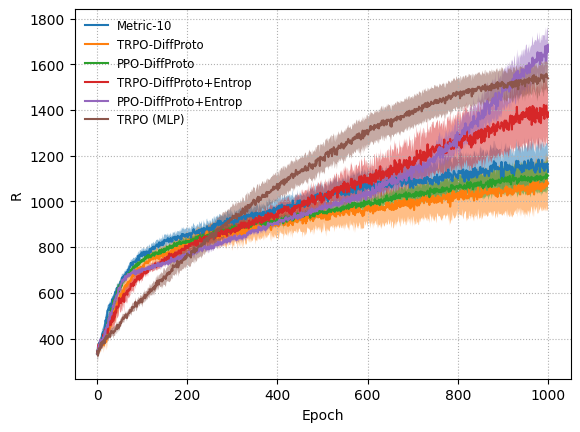}
		\caption{AntBulletEnv-v0}
	\end{subfigure}
	\caption{Performance of TRPO and PPO using a differentiable version of a mixture of expert policy, with and without an entropy constraint, on four \texttt{pybullet} locomotion tasks. We plot our algorithm and TRPO with a neural network policy as reference. All plots averaged over 24 runs, showing mean and 95\% confidence interval.}
	\label{fig:perf_diff}
	\vspace*{-.15in}
\end{figure*}

\begin{figure*}
	\centering
	\begin{subfigure}[t]{\sizechetah\textwidth}
		\includegraphics[width=\textwidth]{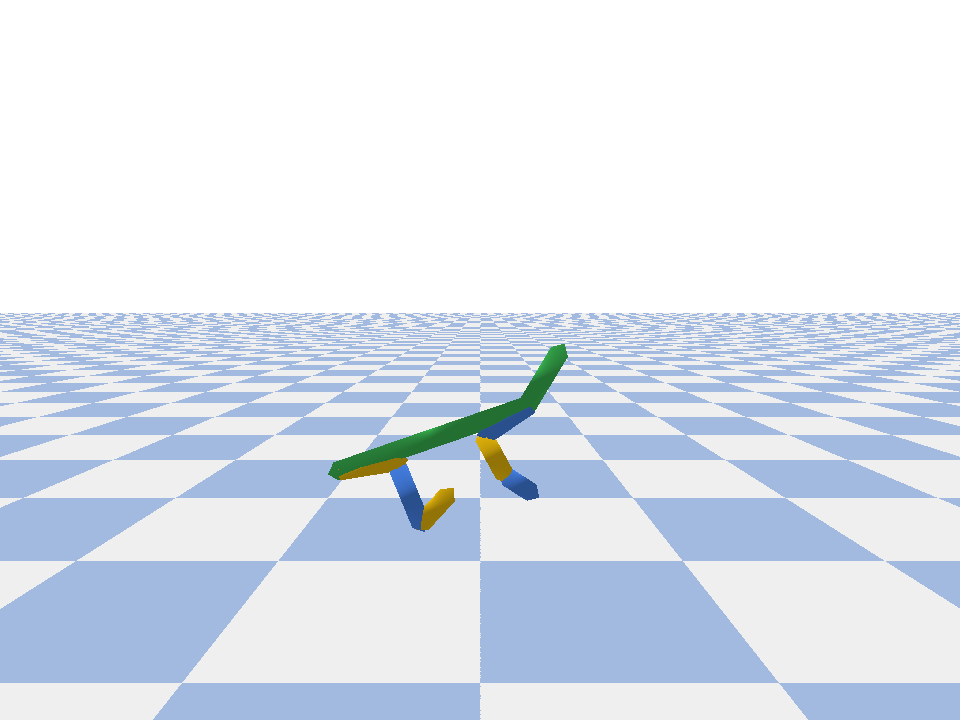}
		\caption{Cluster 1}
	\end{subfigure}
	\hfill
	\begin{subfigure}[t]{\sizechetah\textwidth}
		\includegraphics[width=\textwidth]{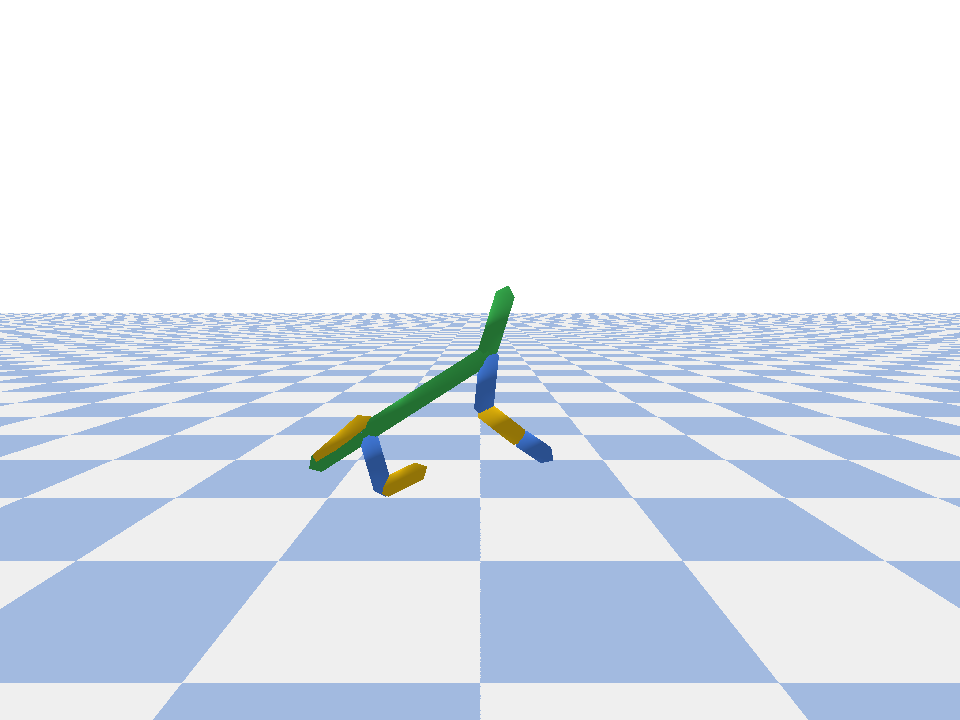}
		\caption{Cluster 2}
	\end{subfigure}
	\hfill
	\begin{subfigure}[t]{\sizechetah\textwidth}
		\includegraphics[width=\textwidth]{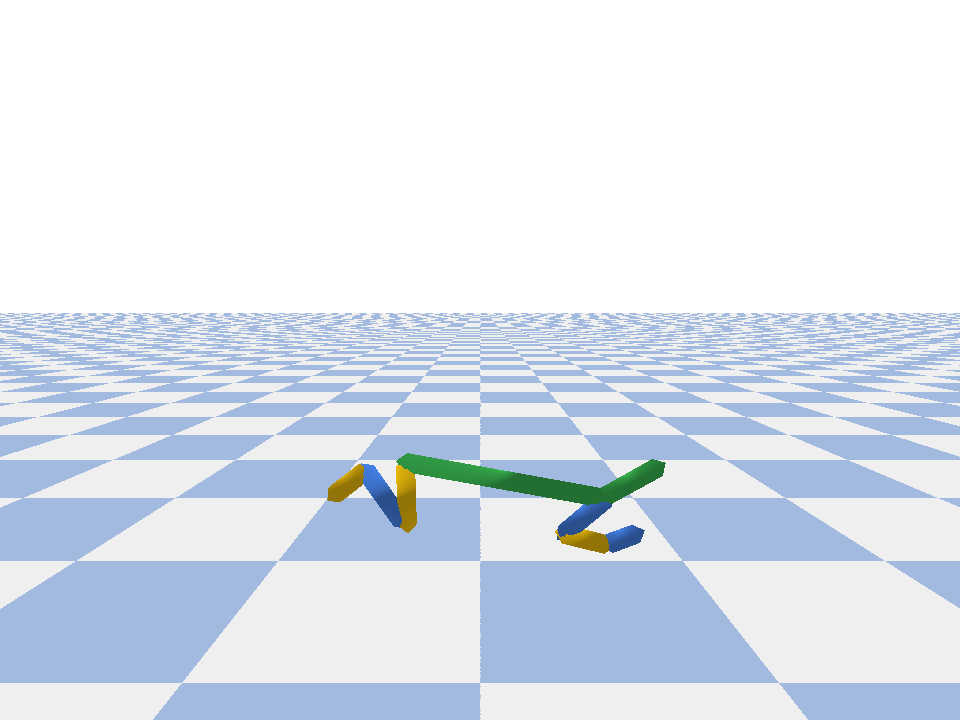}
		\caption{Cluster 3}
	\end{subfigure}
	\hfill
	\begin{subfigure}[t]{\sizechetah\textwidth}
		\includegraphics[width=\textwidth]{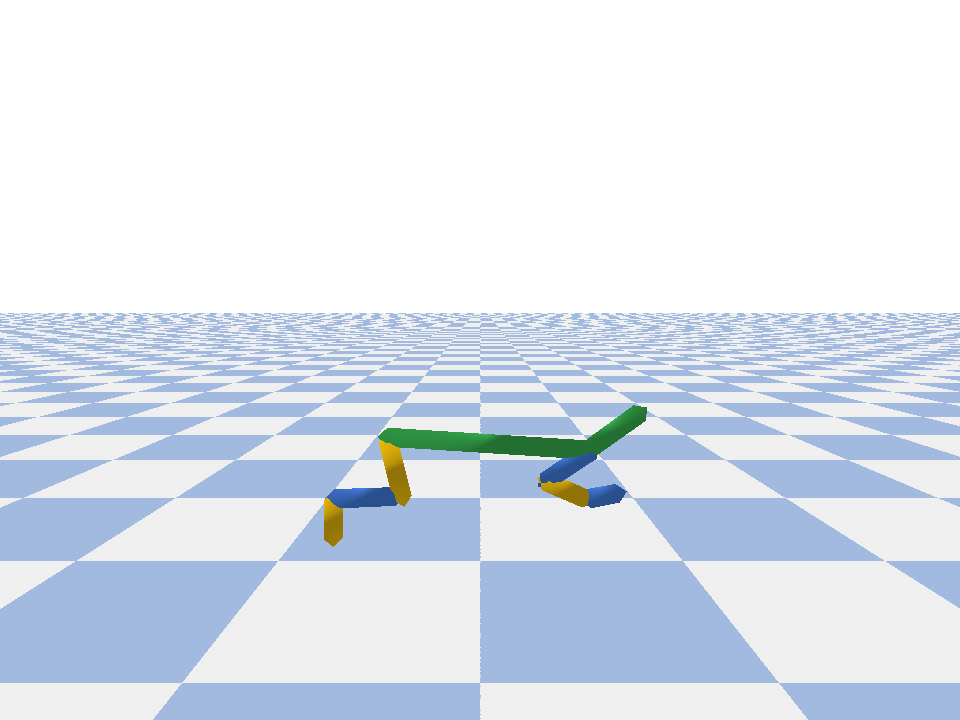}
		\caption{Cluster 7}
	\end{subfigure}
	\begin{subfigure}[t]{\textwidth}
		\centering
		\includegraphics[width=.8\textwidth]{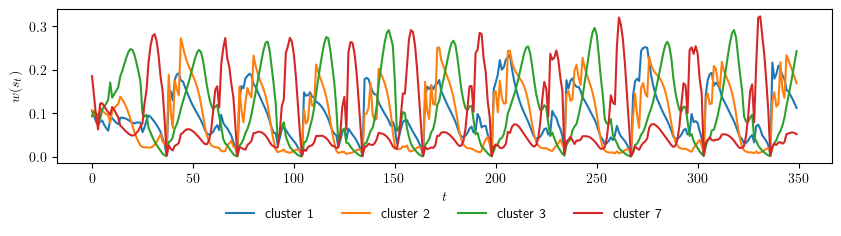}
		\caption{Cluster activation}
		\label{fig:sub:cheetact_diff}
	\end{subfigure}
	\caption{Four clusters out of ten and their associated activation on the \texttt{HalfCheetahBulletEnv-v0} environment learned by PPO and a differentiable version of the mixture of expert policy.}
	\label{fig:cheeact_diff}
\end{figure*}

By visualizing the expert policies, we are able to understand how the movement decomposes into shorter sub-policies. For instance, for the Hopper it becomes clear that Cluster 4 is responsible for the landing, Cluster 0 the propelling and Cluster 9 rotates the ankle's joint right after the propelling and prepares for the landing by having the foot parallel to the floor. Cluster 0 (propelling) is executed in a short burst, and Cluster 4 (landing) is executed the longest as the Hopper spends most of its time in the air. And the motion repeats cyclically, by chaining the propelling, the rotation of the ankle and the landing. Similarly for the HalfCheetah, by inspecting the expert policies and the sequence with which they are activated, we have a clear understanding of how the gait decomposes. First, the propelling is initiated by the leftmost leg (Cluster 9) then the right most one (Cluster 0) and then the landing operates in the opposite order by first extending the rightmost leg~(Cluster 4) and slightly thereafter the leftmost leg (Cluster 2), before repeating the cycle. A similar inspection of the expert policies for the Ant problem gives a clear understanding of how the motion can be decomposed and is deferred to the appendix and supplementary material.

So far we have shown that using our algorithm, elements of the learned policy can be inspected and understood individually. Together with their activation sequence, one can form a good understanding of the decomposition of the movement into specialized skills for each robot's gait. However, one can question whether obtaining these results is due to design and algorithmic decisions we have taken or that other---and perhaps simpler---approaches can yield similar results. In the introduction, we have emphasized on the necessity of picking the prototypical states from trajectory data in order to ensure that the expert policies are interpretable. This choice has introduced algorithmic complexity due to the non-differentiable nature of the prototype picking operation. It appears sensible to test whether this choice was justified or if similarly interpretable policies can be obtained using standard RL algorithms from the literature. 

\subsubsection{Differentiable prototypical states}
To evaluate the importance of selecting prototypical states from trajectory data, we implement within our Python codebase in MushroomRL~\cite{deramo2020mushroomrl}, another baseline that uses standard RL algorithms such as PPO and TRPO to optimize the mixture of expert policy described in Sec.~\ref{sec:pol}. Crucially, to be applicable these algorithms require the policy to be fully differentiable. As result, instead of selecting the cluster centers from trajectory data using the discrete optimization routine described in Sec.~\ref{sec:discopt}, the centers are learned using the gradient-based update of each algorithm. Fig.~\ref{fig:perf_diff}, with labels \texttt{TRPO-DiffProto} and \texttt{PPO-DiffProto}, shows the performance of such an approach. Unfortunately, upon examination of each individual run, none of the algorithms is able to learn any successful gait in any of the environment, even on HalfCheetah where \texttt{PPO-DiffProto} performs better than our algorithm in average. This highlights the difficulty of finding an appropriate clustering of the state at the same time as the policy is learned.

To address this problem, we add to each baseline algorithm an additional entropy constraint, using the same hyper-parameters as in our setting, following \cite{Akrour19}. The entropy constraint allows to maintain exploration for a longer period of time, which allows both algorithms to find suitable prototypes first and to be able to explore and learn a competent policy in a second phase. This two phase behavior is visible in the performance plots in Fig.~\ref{fig:perf_diff}, with labels \texttt{TRPO-DiffProto+Entrop} and \texttt{PPO-DiffProto+Entrop} on most environments. Compared to our algorithm, either one of the baselines---but not both, save for one case---performs better than our algorithm on all environment, as soon as a good cluster partition is found. This is expected since the differentiable problem is easier than our setting. In terms of interpretability, Fig.~\ref{fig:cheeact_diff} shows the clusters returned by the differentiable optimization. We already notice several cases of self-collision and unnatural orientation indicating that the prototypical states left the state space. 
More strikingly, the videos provided in the supplementary
show a large discrepancy between the expert policies and the overall behavior, such that it is impossible to relate the two and understand and decompose the gait, as we were able to do with 
our algorithm.

\section{Conclusion}
We have proposed in this paper a mixture of experts policy structure based on the proximity to a small set of prototypical states. Crucially, the prototypical states were constrained to come from trajectory data, selected by a discrete optimization routine. We have shown that this constraint greatly increases the interpretability of the expert policies compared to a fully differentiable approach that yields prototypical states laying outside of the state space. From a technical point of view, we proposed a policy update that tackles the mixed continuous and discrete optimization problem, and despite the difficulty of the setting, we have shown that on several continuous action RL benchmarks we are able to approach the performance of deep RL baselines that use neural network policies. 


However, while our contributions take a step towards interpretable \gls{rl}, there are still a few limitations that should be addressed in future work. In this paper, we compute the similarity to the prototypical states using an Euclidean distance, which might be ill-suited for more complex state spaces. Instead, the distance should be learned while ensuring that it remains interpretable. One possible direction is to constrain the distance to respect the underlying dynamics of the \gls{mdp}, keeping the intuitive knowledge that similar states likely occur close to each other. Another direction is to extract a semantic and factored understanding of the state and use a simple similarity measure on this representation. Together with the proposed discrete-continuous optimization scheme of this paper, we believe that such an abstract similarity measure will bring us very close to the automated learning of a policy that is as interpretable as the Rubik's cube example discussed in the introduction. 

\bibliographystyle{IEEEtran}
\bibliography{papers,extra,hrl}
\def\vsppaceaut{-.44in}
\vspace*{\vsppaceaut}
\begin{IEEEbiography}[{\includegraphics[width=1in,height=1.25in,clip,keepaspectratio]{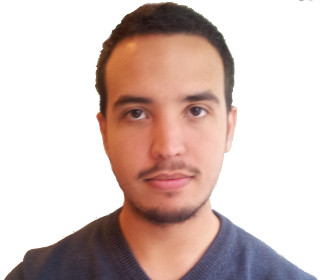}}]{Riad Akrour}
Riad Akrour is a Postdoctoral researcher for the Intelligent Autonomous Systems Laboratory at the Computer Science Department of the Technical University of Darmstadt. Riad Akrour received his M.Sc. degree in Computer Science from Sorbonne University (Paris VI) in 2010 and his Ph.D. in Computer Science from the University of Paris-Saclay (Paris XI) in 2014. His research interests include Reinforcement Learning, Continuous Optimization and Robotics.
\end{IEEEbiography}
\vspace*{\vsppaceaut}
\begin{IEEEbiography}[{\includegraphics[width=1in,height=1.25in,clip,keepaspectratio]{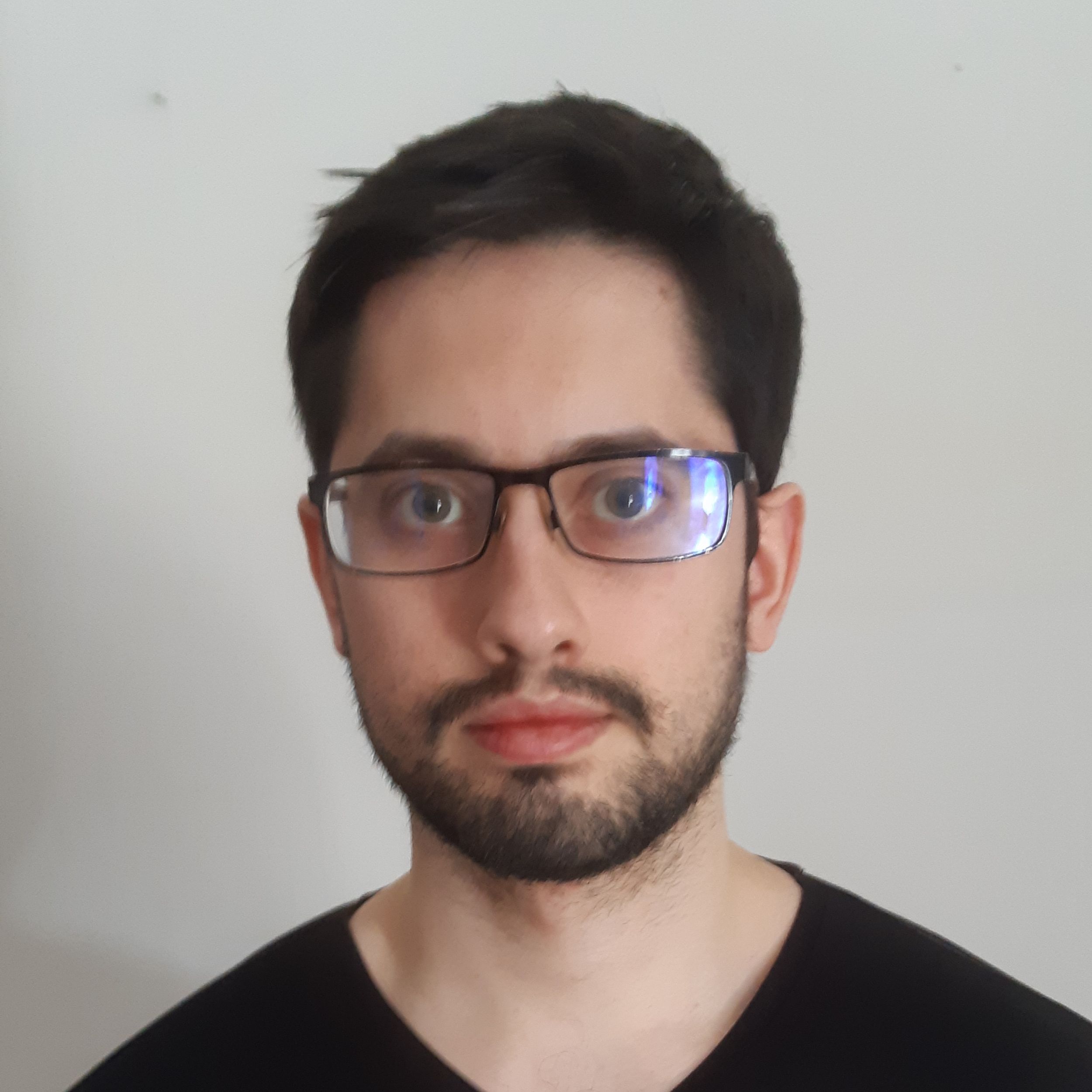}}]{Davide Tateo}
Davide Tateo is a Postdoctoral researcher for the Intelligent Autonomous Systems Laboratory at the Computer Science Department of the Technical University of Darmstadt. Davide Tateo received his M.Sc. degree in Computer Engineering at Politecnico di Milano in 2013 and his Ph.D. in Information Technology from the same university in 2017. His research interests include Deep Reinforcement Learning, Robotics, and Robot Learning, focusing on learning high-speed motion primitives.
\end{IEEEbiography}
\vspace*{\vsppaceaut}
\begin{IEEEbiography}[{\includegraphics[width=1in,height=1.25in,clip,keepaspectratio]{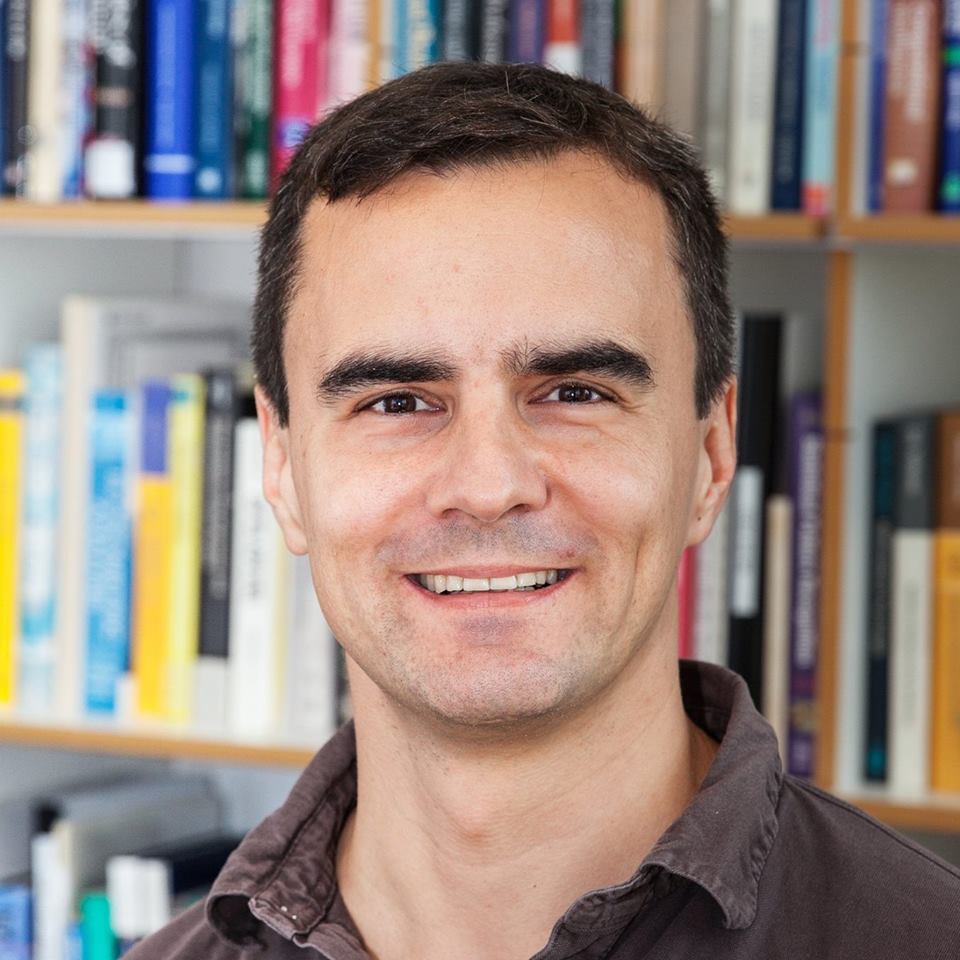}}]{Jan Peters}
Jan Peters is a full professor (W3) for Intelligent Autonomous Systems at the
Computer Science Department of the Technische Universitaet Darmstadt and at the same time a senior research scientist and group leader at the Max-Planck Institute for Intelligent Systems, where he heads the interdepartmental Robot Learning Group. Jan Peters has received the Dick Volz Best 2007 US PhD Thesis Runner-Up Award, the Robotics: Science \& Systems - Early Career Spotlight, the INNS Young Investigator Award, and the IEEE Robotics \& Automation Society's Early Career Award as well as numerous best paper awards. In 2015, he received an ERC Starting Grant and in 2019, he was appointed as an IEEE Fellow.
\end{IEEEbiography}

\end{document}